\documentclass[10pt]{article} 
\usepackage[accepted]{rlj}

\usepackage{amsmath,amsfonts,bm}

\def\eqref#1{equation~\ref{#1}}
\def\Eqref#1{Equation~\ref{#1}}

\def\1{\bm{1}}

\DeclareMathAlphabet{\mathsfit}{\encodingdefault}{\sfdefault}{m}{sl}
\SetMathAlphabet{\mathsfit}{bold}{\encodingdefault}{\sfdefault}{bx}{n}

\newcommand{\Div}{\mathrm{Div}}
\newcommand{\D}{\mathcal{D}}

\DeclareMathOperator*{\argmax}{arg\,max}

\usepackage[utf8]{inputenc} 
\usepackage[T1]{fontenc}    
\usepackage{hyperref}       
\usepackage{url}            
\usepackage{booktabs}       
\usepackage{amsfonts}       
\usepackage{nicefrac}       
\usepackage{microtype}      
\usepackage{xcolor}         
\usepackage{graphicx}
\usepackage{comment}
\usepackage{amsmath,amssymb}
\usepackage{amsthm}
\usepackage{tikz}
\usepackage{dsfont}
\usepackage{wrapfig}
\usepackage{enumitem}
\usetikzlibrary{positioning, calc, backgrounds}

\newtheorem{theorem}{Theorem}

\title{Minimal Ingredients for Reward Assignment from \\Expert Demonstrations}

\setrunningtitle{Minimal Ingredients for Reward Assignment from Expert Demonstrations}

\author{
Zixuan Dong\textsuperscript{1,3,$\dagger$},
Yumi Omori\textsuperscript{2,$\dagger$},
Keith Ross\textsuperscript{2,$\ast$}
}

\emails{
\{zixuandong, yd2247, keithwross\}@nyu.edu
}

\affiliations{
\textsuperscript{1}\textbf{New York University} \\
\textsuperscript{2}\textbf{New York University Abu Dhabi} \\
\textsuperscript{3}\textbf{SFSC of AI and DL, NYU Shanghai}
\par
\textsuperscript{$\dagger$}Equal contribution
\textsuperscript{$\ast$}Corresponding author
}

\contribution{
A systematic comparison of proximity-based reward assignment along two design axes: proximity approximation and temporal correspondence. This examines the structural inductive biases underlying many existing methods.
}
{
Prior work typically introduces independent algorithms evaluated as monolithic systems, making it difficult to understand which structural ingredients drive performance.
}

\contribution{
An extensive empirical study across 32 benchmarks spanning offline and online RL, multi-expert regimes, and four downstream learners (IQL, ReBRAC, DrQ-v2, TD3+BC). We find that proximity alone provides a strong signal for offline RL, while lightweight temporal correspondence yields modest gains offline but becomes essential online or with multiple demonstrations.
}
{
Most prior work claims performance improvements over oracle RL learners but evaluates on limited settings or a single downstream algorithm, leaving robustness across learners and offline–online regimes underexplored.
}

\contribution{
A lightweight theoretical result helps characterize when minimalist proximity-based rewards suffice for effective offline RL, compared with Optimal Transport approaches.
}
{
This result is motivated by \citet{yu2022uds} and our proof incorporates their proof of Theorem 4.1. By adapting to our context, we better understand some empirical results.
}

\summary{

Reward assignment from scarce demonstrations is a key challenge in both offline and online imitation learning. A common and intuitive strategy assigns rewards according to how closely learner trajectories match expert demonstrations. Although this principle underlies many existing methods, including recent approaches based on Optimal Transport (OT), the specific contributions of their core ingredients remain systematically underexplored. We therefore ask: what is the minimal structure a reward assignment rule must encode to achieve strong downstream RL performance across settings? We approach this question along two design axes: \textit{proximity approximation} and \textit{temporal alignment}. Along the first axis, we compare the Wasserstein distance with a simple point-to-set distance. Along the second, we impose temporal constraints under the assumption that the learning agent moves at a speed comparable to the expert. Across 32 benchmarks spanning offline and online settings, our key empirical findings are: (1) In offline regimes, gains from complex approximation and temporal constraints largely diminish with a stronger, well-tuned RL learner; (2) However, with an increasing number of expert demonstrations, temporal alignment becomes more important to prevent substantial performance degradation for offline RL learners; and (3) In online regimes, while sophisticated proximity approximation remains secondary, the absence of temporal constraints consistently entails little or no learning progress. We further complement our offline results with a lightweight theoretical analysis characterizing when a point-to-set distance suffices. Overall, these findings advocate algorithmic minimalism in reward design before introducing complex schemes in both offline and online imitation learning.
}


\begin{document}

\makeCover  
\maketitle  

\begin{abstract}
    Reward assignment from scarce demonstrations is a key challenge in both offline and online imitation learning. A common and intuitive strategy assigns rewards according to how closely learner trajectories match expert demonstrations. Although this principle underlies many existing methods, the core ingredients that drive performance remain systematically underexplored. We therefore ask: what is the minimal structure that reward assignment must encode to achieve effective downstream RL performance across settings? We approach this question along two design axes: \textit{proximity approximation} and \textit{temporal alignment}. Across 32 benchmarks spanning offline and online settings, and with three downstream RL algorithms, our empirical findings suggest: (1) In offline regimes, proximity alone captures the reward structure necessary for effective offline RL, while (2) lightweight temporal correspondence provides consistent gains that are modest offline but essential online or in the presence of multiple demonstrations. We further complement our offline results with a lightweight theory characterizing when simple proximity approximation suffices. Overall, these findings advocate algorithmic minimalism in reward design before introducing complex schemes in both offline and online imitation learning. Our code is available at \href{https://github.com/Konokiii/Minimalist_RL_Reward}{Minimalist\_RL\_Reward}.
\end{abstract}

\section{Introduction}

Reward assignment from demonstrations lies at the core of both offline and online imitation learning. When task rewards are unavailable, learning performance depends almost entirely on the quality of the proxy reward signal constructed from expert data, a setting closely related to inverse reinforcement learning (IRL) \citep{Ng2000irl_algos}. In practice, one often has access to only a small number of demonstrations, and the central challenge is therefore to convert these demonstrations into a reward function that reliably drives downstream reinforcement learning.

A large body of recent work addresses this challenge by using demonstrations to relabel trajectories with rewards derived from matching: states or trajectory segments that resemble the expert are rewarded, whereas those that deviate are penalized. This strategy is particularly appealing when demonstrations are scarce, e.g., one single trajectory, where supervised imitation can become brittle and where proximity to expert behavior provides a natural learning signal. Within this family of approaches, Optimal Transport (OT) methods have emerged as a prominent class \citep{dadashi2021pwil, haldar2022watch_match,haldar2023teach_fish,ot_reward}. These methods compute a transport plan between learner and expert trajectories and convert the induced alignment into rewards; temporally structured variants further impose progress constraints intended to respect the sequential nature of control \citep{ temporal_ot, huey2025orca}.

Despite strong empirical results, it remains unclear which design choices in these matching-based reward constructions are actually essential. \textbf{When proxy rewards are ultimately computed from distances between learner and expert states, what is the minimal structural content that a reward assignment rule must encode in order to achieve effective downstream RL performance across settings?}

In this paper, we address this question by examining two design axes that the literature has argued are critical for proximity-based reward relabeling. The first axis concerns proximity approximation: how the distance between a state and a demonstration trajectory is measured and aggregated. The second axis concerns temporal alignment: whether the notion of closeness is invariant to state reordering, or instead depends on respecting progress along the expert trajectory. Along the first axis, we compare OT-based Wasserstein distance, as instantiated in reward assignment methods \citep{ot_reward,temporal_ot}, with a simple point-to-set distance. Along the second axis, we follow a common assumption: when imposing temporal constraints on reward labeling, the learning agent is assumed to move at a speed comparable to the expert \citep{liu2024ads,temporal_ot}.

To instantiate this comparison concretely, we consider four reward assignment rules spanning these axes in both offline and online learning pipelines: \textsc{OT} \citep{ot_reward} and a temporally constrained OT variant (\textsc{TemporalOT}) \citep{temporal_ot} as representatives of coupling approaches, alongside two minimalist rules that isolate the same structural ingredients without solving a transport problem: minimum distance (\textsc{Min-Dist}) and a simple progress-based segment correspondence rule (\textsc{Seg-Match}). This controlled setup allows us to assess when sophisticated proximity computations or temporal constraints are genuinely necessary, and when they primarily repackage simpler inductive biases. To ensure our conclusions are not artifacts of a particular downstream RL learner, we evaluate across multiple RL algorithms, including IQL \citep{iql}, ReBRAC \citep{tarasov2023rebrac}, and DrQ-v2 \citep{drqv2}, and we additionally evaluate TD3+BC \citep{td3_bc} in Appendix~\ref{para:td3+bc}.

Across extensive experiments on 32 benchmarks, complemented by lightweight theoretical analysis, our results present a consistent picture:
\begin{itemize}
    \item In offline learning, proximity alone often provides a strong signal, and more complex proximity computations or temporal constraints are frequently unnecessary.
    \item As the number of demonstrations increases, however, temporal structure becomes more important: progress dependence stabilizes learning and prevents substantial performance degradation arising from ambiguity in purely proximity-based matching across multiple experts.
    \item In the online setting, a different failure mode dominates. When the replay buffer is populated through exploration and credit assignment is fragile, enforcing temporal progress is essential for producing a usable learning signal, whereas the specific proximity approximation plays a comparatively minor role. In this regime, a simple progress-based rule can be competitive with temporally constrained OT while avoiding global coupling optimization.
\end{itemize}

Taken together, these findings suggest that future reward assignment algorithms should prioritize simple structural ingredients before adopting more complex schemes in both offline and online imitation learning.

\section{Preliminary}
\label{sec:prelim}
An episodic Markov Decision Process (MDP) can be represented as a tuple:
\[
\mathcal{M} = (\mathcal{S}, \mathcal{A}, R, P, \rho_0, \gamma, T)
\]
where $\mathcal{S}$ and $\mathcal{A}$ are the state and action spaces; $R: \mathcal{S} \times \mathcal{A} \rightarrow \mathbb{R}$ is the reward function; $P: \mathcal{S} \times \mathcal{A} \rightarrow \Delta(\mathcal{S})$ is the transition function; $\rho_0$ is the initial state distribution; $\gamma \in (0,1]$ is the discount factor; and $T$ is the maximal episode length. 

Denote a reward-free trajectory as $\tau = (s_0, a_0, s_1, a_1, \dots, s_T, a_T)$. We consider a setting where we are given $K$ expert demonstrations $\mathcal{D}_e = \{\tau_k^e\}_{k=1}^K$ and where $K$ is small, e.g., $K=1$. Our focus is to (1) assign proxy rewards $\hat r$ from demonstrations to arbitrary reward-free trajectories $\{\tau_n\}_{n=1}^N$, and (2) then train a policy by a standard RL learner with the relabeled data $\{(s_0, a_0, \hat r_0, \dots, s_{T_n}, a_{T_n}, \hat r_{T_n})_n\}_{n=1}^N$.
We study both offline and online pipelines:
In offline settings, the learner is trained from a fixed dataset of trajectories without task rewards;
In online settings, the learner interacts with the environment while receiving proxy rewards computed from the expert data.

\paragraph{Proximity-based reward assignment}
Proxy rewards $\hat r$ can often be derived from the proximity between agent states and expert states \citep{ot_reward, liu2023clue, lyu2024seabo, temporal_ot}.
Given an arbitrary trajectory $\tau = (s_1, \dots, s_T)$ and an expert trajectory $\tau^e = (s_1^e, \dots, s_{T_e}^e)$, define a cost matrix $C_{\tau,\tau^e} \in \mathbb{R}_+^{T \times T_e}$ with $C_{\tau,\tau^e}^{ij} = c(s_i, s_j^e)$ where $c(s,s^e)\geq 0$ measures the distance between states. A generalized construction then assigns a proximity-based reward to an agent state $s_i$ by weighting the negative costs across expert states:
\begin{equation}
\hat r(s_i) \;=\; - \sum_{j=1}^{T_e} w_{ij} \, c(s_i, s_j^e)
\label{eq:alignment_reward}
\end{equation}

This viewpoint is helpful in our analysis because it makes explicit the two structural ingredients that shape the rewards:
\emph{(1) how weights $w_{ij}$ are calculated to approximate proximity to an expert trajectory}, and \emph{(2) how those weights incorporate temporal constraints to prevent proxy rewards from being invariant to state reordering}.

\subsection{Optimal Transport Reward}
\label{sec:otr}
A representative instantiation along the first design axis computes $w_{ij}$ using entropy-regularized OT \citep{cuturi2013sinkhorn}. The algorithm computes a minimal-cost coupling between the non-expert and expert trajectories by optimizing the following Wasserstein distance objective:
\[
\mathcal{W}(\tau, \tau^e) = \min_{\mu \in \mathbb{R}_+^{T \times T_e}} \langle \mu, C \rangle_F \;-\; \varepsilon \, \mathcal{H}(\mu)
\quad
\text{s.t.}\quad
\mu \mathbf{1} = \tfrac{1}{T}\mathbf{1}, \ \mu^\top \mathbf{1} = \tfrac{1}{T_e}\mathbf{1}
\]
where \( \mu \) is a coupling matrix that specifies the transport of probability mass between two trajectories; $\langle \cdot, \cdot\rangle_F$ is the Frobenius inner product; $\mathcal{H}$ is the entropy; and $\varepsilon$ controls the regularization strength. Let \( \mu^* \) denote the optimal transport plan that minimizes the above objective. The OT reward \citep{ot_reward} then sets $w_{ij} = \mu_{ij}^*$ in \Eqref{eq:alignment_reward} to assign proximity-based rewards to unlabeled trajectories.

\subsection{Temporal Optimal Transport Reward}
\label{sec:temporalot}
Prior work \citep{temporal_ot} has identified two drawbacks of the OT reward-labeling approach. First, the OT solution is \emph{temporally invariant}: because the Wasserstein distance treats trajectories as unordered sets of states, trajectories that differ only in temporal ordering receive identical rewards. Second, OT-based rewards are \emph{non-stationary}, as the reward assigned to a state depends on the entire trajectory, including temporally distant states, which causes identical state transitions to receive different rewards.

TemporalOT \citep{temporal_ot} then adds to the second design choice by injecting temporal structure via two modifications. First, it replaces the pairwise cost with a context-aware cost:
\begin{equation*}
\tilde c(s_i, s_j^e) \;=\; \frac{1}{k_c} \sum_{h=0}^{k_c-1} c(s_{i+h}, s_{j+h}^e)
\label{eq:context_cost}
\end{equation*}
where a context length $k_c$ smooths distance estimates using short local neighborhoods.
Second, it constrains coupling matrices to be temporally local through a mask $M$ that restricts feasible couplings to indices within a window of radius $k_m$ around the diagonal. That is, $M_{i,j} = 1, \text{if } j \in [i - k_m, i + k_m]$ and $M_{i,j}=0$ otherwise. In matrix form, TemporalOT then solves the following problem:
\begin{align*}
\mathcal{W}(\tau, \tau^e) = \min_{\mu \in \mathbb{R}_+^{T \times T_e}} \left\langle M \odot \mu, \tilde{C} \right\rangle_F - \varepsilon \mathcal{H}(M \odot \mu) \quad \text{s.t.}\quad (M \odot \mu) \mathbf{1} = \frac{1}{T} \mathbf{1}, \; (M \odot \mu)^\top \mathbf{1} = \frac{1}{T_e} \mathbf{1}
\end{align*}
where $\odot$ is the Hadamard product operator.

\section{Methods: Simple Reward Assignment Rules}
\label{sec:simple}
Proximity-based rewards in~\Eqref{eq:alignment_reward} can be viewed as assigning each agent state $s_t$ a proxy reward by comparing it to expert states using a distance measure $\mathrm{Dist}(\cdot,\cdot)$ and a temporal correspondence rule that determines which expert states are relevant. In this section, we define two simple reward assignment rules that instantiate the two design axes discussed. The first approximates proximity by replacing the OT-based Wasserstein distance with a simple point-to-set distance. The second captures temporal alignment by enforcing progress along the expert trajectory under the same assumption as TemporalOT: that the agent moves at a speed comparable to the expert \citep{liu2024ads, temporal_ot}. These constructions provide minimalist alternatives to OT and TemporalOT rewards and help interpret the empirical findings observed in the next section.

\subsection{Minimum-distance Reward}
\label{sec:mindist}
For a non-expert state $s_t$ and an expert demonstration $\tau^e$, we define the Min-Dist reward as the negative distance to the closest expert state:
\begin{equation*}
\hat r_{\mathrm{min}}(s_t)
\;:=\;
- \min_{s^e \in \tau^e} \mathrm{Dist}(s_t, s^e)
\label{eq:mindist}
\end{equation*}
where $\mathrm{Dist}(\cdot,\cdot)$ can be any distance measure, including Euclidean distance and cosine distance.
This rule assigns high rewards to states that are close to any expert state and ignores temporal ordering or the phase of the trajectory.

\subsection{Segment-matching Reward}
\label{sec:segmatch}
The second rule augments the point-to-set distance with a lightweight temporal correspondence. Given a non-expert trajectory $\tau$ of length $T$ and an expert demonstration $\tau^e$ of length $T_e$, we partition the expert states into $T$ consecutive, non-overlapping segments $\{\Gamma_1,\dots,\Gamma_T\}$ as evenly as possible, and associate timestep $t$ of the non-expert trajectory with segment $\Gamma_t$.
Let $q = \left\lfloor \tfrac{T_e}{T} \right\rfloor$ and $l = T_e \bmod T$. Formally, each segment is defined by
\begin{equation}
\Gamma_t
=
\left\{ s_i^e \ \middle|\ i \in [a_t,\, b_t] \cap \mathbb{Z} \right\}
\label{eq:segments}
\end{equation}
where
\[
a_t = (t-1)q + 1 + \min(t-1,l),
\qquad
b_t = tq + \min(t,l)
\]

When $T_e$ is a multiple of $T$, each segment contains the same number of expert states; when $T_e = T$, each segment reduces to a single expert state.

For each timestep $t$, we then assign the Seg-Match reward as the negative distance to the closest expert state within the corresponding segment $\Gamma_t$:
\begin{equation*}
\hat r_{\mathrm{seg}}(s_t)
\;:=\;
- \min_{s^e \in \Gamma_t} \mathrm{Dist}(s_t, s^e).
\label{eq:segmatch}
\end{equation*}
This rule encodes a mild assumption that trajectories can be compared by coarse phase, in the sense that states occurring at similar relative progress along the trajectory should be compared to one another \citep{liu2024ads, temporal_ot}.
Unlike OT-based temporal constraints, this correspondence is fixed and does not require solving for an optimal coupling.

When the expert demonstration is shorter than the non-expert trajectory, i.e., $T > T_e$, the segment construction in~\Eqref{eq:segments} yields empty segments for $t > T_e$. In this case, we assign rewards based on distance to the final expert state:
\begin{equation}
\hat r_{\mathrm{seg}}(s_t)
\;:=\;
- \mathrm{Dist}(s_t, s_{T_e}^e),
\qquad t > T_e
\label{eq:segmatch_tail}
\end{equation}

In Appendix~\ref{appx:generalized_reward}, we comment on the computational complexity of the four methods introduced so far. Assuming state dimension $d$ and $T \approx T_e$, we summarize costs for labeling one trajectory in Table~\ref{tabel:complexity}.

\begin{table}[htb]
\caption{Computational complexity of different reward labeling methods. We assume $T \approx T_e$. Min-Dist uses KD tree nearest neighbor search. OT based methods additionally require iterative transport solvers.}
\label{tabel:complexity}
\centering
\scalebox{0.9}{
\begin{tabular}{lcccc}
\toprule
\textbf{Algorithms} & OT & TemporalOT & Seg-Match & Min-Dist \\
\hline
\textbf{Complexity} &
$\mathcal{O}(dT^2)$ &
$\mathcal{O}\!\left((d+k_c)T^2\right)$ &
$\mathcal{O}(dT)$ &
$\mathcal{O}\!\left( (\log T + d)T\right)$ \\
\bottomrule
\end{tabular}}
\end{table}

\section{Experiments}
\label{sec:exps}
We evaluate the four reward assignment methods that instantiate the two design axes across both offline and online RL settings. Our main comparisons include OT reward \citep{ot_reward}, its temporally constrained variant TemporalOT \citep{temporal_ot}, and the two minimalist reward assignment rules defined in Section~\ref{sec:simple}. We also include an Oracle baseline, where the downstream learner is trained with ground truth task rewards. To test whether conclusions depend on the downstream learner, we report results with multiple RL algorithms, including IQL, ReBRAC, and DrQ-v2. And in the offline setting, we report behavior cloning (BC) as an additional reference. Throughout, proxy rewards are calculated using cosine distance following the literature \citep{dadashi2021pwil, ot_reward, temporal_ot}.

We first report offline results on D4RL with two offline RL learners in Section~\ref{sec:exps_offline_rl}. We then consider online RL in Section~\ref{sec:exps_online_rl}. Section~\ref{sec:multi-expert-main} studies scaling with additional expert demonstrations. Finally, Section~\ref{sec:thm} provides a theoretical result that helps interpret the empirical trends.

\subsection{Offline RL Evaluation}
\label{sec:exps_offline_rl}

\paragraph{Setup}
We evaluate on D4RL \citep{d4rl}, covering \texttt{MuJoCo-v2}, \texttt{Adroit-v1}, and \texttt{Antmaze-v2}, for a total of 23 datasets.
Following \citet{ot_reward}, for each dataset we first select a single expert trajectory as the highest-return trajectory in the original D4RL dataset, and then remove ground-truth rewards from all transitions to form an unlabeled offline dataset $\mathcal{D}_o$.
This uses task reward only to choose a fixed demonstration; reward labeling and downstream training use no task reward.
Each reward assignment method then produces proxy rewards for all transitions in $\mathcal{D}_o$, after which we train an offline RL policy.

Similar to prior work \citep{dadashi2021pwil, ot_reward}, we post-process relabeled rewards, first with an exponential squashing function and then rescaling with the global dataset statistics. For OT and TemporalOT, we use the reported squashing parameters from \citet{ot_reward}; for our two minimalist rules, we use a single default squashing setting without tuning. We provide full details in Appendix~\ref{appx:offline_postprocessing}.

\paragraph{IQL Learner}
We first use Implicit Q-Learning \citep{iql} as the downstream learner, matching the pipeline widely considered in the related work \citep{liu2023clue, ot_reward, lyu2024seabo, bobrin2024align_intents, lyu2025crossdomain_ot}.
We use the same codebase as \citet{ot_reward} and keep hyperparameters fixed across all reward assignment methods.
The only implementation change is to enable state normalization for \texttt{MuJoCo} and \texttt{Adroit}, applied uniformly across all methods, including Oracle, which we found improves training stability and overall performance.
Appendix~\ref{appx:hyperparam_offline} summarizes the full offline setup.

Table~\ref{tabel:iql_main} reports D4RL normalized scores with standard deviations over 10 seeds.
We report the average over the last four evaluations rather than the final evaluation to reduce sensitivity to late training fluctuations \citep{offline_synthetic_pretraining}.
We highlight scores that are within 95\% of the best in each row as a visual aid, excluding Oracle and BC, which are shown for reference\footnote{BC performances are copied from Table 1 provided by \citet{corl_codebase}}.

\begin{table}[ht]
    \caption{Normalized scores with IQL as downstream RL algorithm. }
    \label{tabel:iql_main}
    \centering
    \scalebox{0.7}
    {
    \begin{tabular}{lccccccccccccc}

\toprule
Dataset & BC & IQL (oracle) & OT & TemporalOT & Seg-match & Min-Dist \\
\hline
hopper-medium-replay & 29.81 $\pm$ 2.07 & 94.66 $\pm$ 8.2 & 65.59 $\pm$ 20.5 & 77.01 $\pm$ 4.9 & \textbf{87.69} $\pm$ 1.1 & \textbf{86.96} $\pm$ 2.1 \\
hopper-medium & 53.51 $\pm$ 1.76 & 63.50 $\pm$ 4.8 & \textbf{73.68} $\pm$ 4.8 & \textbf{76.30} $\pm$ 4.9 & \textbf{76.77} $\pm$ 4.5 & 72.39 $\pm$ 4.2 \\
hopper-medium-expert & 52.30 $\pm$ 4.01 & 103.67 $\pm$ 8.9 & \textbf{105.45} $\pm$ 7.6 & \textbf{104.51} $\pm$ 11.5 & \textbf{107.85} $\pm$ 4.0 & 82.67 $\pm$ 37.7 \\
halfcheetah-medium-replay & 35.66 $\pm$ 2.33 & 43.12 $\pm$ 1.6 & \textbf{41.53} $\pm$ 0.4 & \textbf{43.15} $\pm$ 0.6 & \textbf{41.03} $\pm$ 1.0 & \textbf{42.38} $\pm$ 0.9 \\
halfcheetah-medium & 42.40 $\pm$ 0.19 & 47.49 $\pm$ 0.3 & \textbf{43.46} $\pm$ 0.3 & \textbf{45.13} $\pm$ 0.2 & \textbf{43.23} $\pm$ 0.3 & \textbf{45.03} $\pm$ 0.2 \\
halfcheetah-medium-expert & 55.95 $\pm$ 7.35 & 91.24 $\pm$ 2.3 & \textbf{88.53} $\pm$ 3.4 & \textbf{89.49} $\pm$ 3.5 & \textbf{90.71} $\pm$ 2.7 & \textbf{88.72} $\pm$ 2.9 \\
walker2d-medium-replay & 21.80 $\pm$ 10.15 & 73.79 $\pm$ 8.9 & 53.87 $\pm$ 19.5 & \textbf{69.73} $\pm$ 11.4 & \textbf{70.93} $\pm$ 14.6 & \textbf{70.82} $\pm$ 8.6 \\
walker2d-medium & 63.23 $\pm$ 16.24 & 80.57 $\pm$ 4.2 & \textbf{79.58} $\pm$ 1.6 & \textbf{79.83} $\pm$ 1.8 & \textbf{80.20} $\pm$ 2.4 & \textbf{79.43} $\pm$ 1.1 \\
walker2d-medium-expert & 98.96 $\pm$ 15.98 & 111.06 $\pm$ 0.2 & \textbf{109.93} $\pm$ 0.2 & \textbf{110.19} $\pm$ 0.2 & \textbf{110.31} $\pm$ 0.2 & \textbf{110.91} $\pm$ 0.2 \\
\bottomrule
\textbf{MuJoCo-total}& 453.6 & 709.10 & 661.62 & \textbf{695.34} & \textbf{708.73} & \textbf{679.33} \\
\bottomrule
\toprule
pen-human & 71.03 $\pm$ 6.26 & 69.86 $\pm$ 17.2 & 68.20 $\pm$ 17.3 & \textbf{71.43} $\pm$ 20.8 & \textbf{72.28} $\pm$ 17.8 & 66.73 $\pm$ 19.1 \\
pen-cloned & 51.92 $\pm$ 15.15 & 70.96 $\pm$ 20.3 & \textbf{61.50} $\pm$ 18.8 & \textbf{62.13} $\pm$ 19.5 & 54.78 $\pm$ 17.5 & 55.97 $\pm$ 20.1 \\
door-human & 2.34 $\pm$ 4.00 & 3.34 $\pm$ 1.6 & \textbf{3.26} $\pm$ 1.8 & 3.07 $\pm$ 1.7 & 3.10 $\pm$ 1.8 & 2.87 $\pm$ 1.5 \\
door-cloned & -0.09 $\pm$ 0.03 & 0.45 $\pm$ 1.0 & \textbf{1.28} $\pm$ 1.4 & 0.43 $\pm$ 0.7 & 0.46 $\pm$ 1.0 & 0.62 $\pm$ 1.0 \\
hammer-human & 3.03 $\pm$ 3.39 & 1.78 $\pm$ 0.4 & 1.82 $\pm$ 0.4 & \textbf{1.97} $\pm$ 0.8 & \textbf{1.88} $\pm$ 1.3 & 1.77 $\pm$ 0.5 \\
hammer-cloned & 0.55 $\pm$ 0.16 & 1.30 $\pm$ 0.7 & \textbf{0.92} $\pm$ 0.7 & 0.79 $\pm$ 0.3 & \textbf{0.92} $\pm$ 0.7 & 0.87 $\pm$ 0.5 \\
relocate-human & 0.04 $\pm$ 0.03 & 0.19 $\pm$ 0.3 & 0.14 $\pm$ 0.2 & 0.09 $\pm$ 0.1 & \textbf{0.20} $\pm$ 0.3 & 0.10 $\pm$ 0.1 \\
relocate-cloned & -0.06 $\pm$ 0.01 & -0.03 $\pm$ 0.1 & 0.01 $\pm$ 0.1 & -0.02 $\pm$ 0.1 & 0.03 $\pm$ 0.1 & \textbf{0.04} $\pm$ 0.1 \\
\bottomrule
\textbf{Adroit-total} & 128.76 & 147.85 & \textbf{137.14} & \textbf{139.89} & \textbf{133.65} & 128.97 \\
\bottomrule
\toprule
antmaze-umaze-play & 55.25 $\pm$ 4.15 & 86.35 $\pm$ 3.5 & 81.30 $\pm$ 5.4 & 82.05 $\pm$ 4.0 & 78.10 $\pm$ 5.3 & \textbf{88.98} $\pm$ 2.8 \\
antmaze-umaze-diverse & 47.25 $\pm$ 4.09 & 64.05 $\pm$ 8.1 & 66.45 $\pm$ 6.6 & 60.62 $\pm$ 8.1 & 68.00 $\pm$ 11.7 & \textbf{71.60} $\pm$ 7.8 \\
antmaze-medium-play & 0.00 $\pm$ 0.00 & 75.45 $\pm$ 4.7 & \textbf{75.03} $\pm$ 4.8 & \textbf{72.75} $\pm$ 4.6 & 69.83 $\pm$ 4.3 & \textbf{73.50} $\pm$ 4.8 \\
antmaze-medium-diverse & 0.75 $\pm$ 0.83 & 72.92 $\pm$ 5.4 & \textbf{72.53} $\pm$ 5.3 & \textbf{73.05} $\pm$ 3.7 & \textbf{71.80} $\pm$ 5.2 & \textbf{70.98} $\pm$ 4.0 \\
antmaze-large-play & 0.00 $\pm$ 0.00 & 46.38 $\pm$ 5.5 & 44.93 $\pm$ 5.2 & 45.67 $\pm$ 6.5 & \textbf{49.58} $\pm$ 6.1 & \textbf{48.03} $\pm$ 6.2 \\
antmaze-large-diverse & 0.00 $\pm$ 0.00 & 45.15 $\pm$ 5.4 & 44.92 $\pm$ 5.0 & \textbf{48.58} $\pm$ 5.4 & 46.62 $\pm$ 7.0 & \textbf{50.78} $\pm$ 6.3 \\
\bottomrule
\textbf{Antmaze-total}& 103.26 & 390.30 & \textbf{385.15} & 382.73 & \textbf{383.93} & \textbf{403.85} \\
\bottomrule
     
    \end{tabular}
    }
    
\end{table}

Three patterns stand out in Table~\ref{tabel:iql_main}.
First, adding coarse temporal correspondence consistently improves reward assignment in locomotion style domains.
Methods that encode temporal structure, such as TemporalOT and Seg-Match, are typically strongest on \texttt{MuJoCo}.
Second, nearest-neighbor proximity alone already provides a competitive learning signal in several settings.
The Min-Dist reward performs strongly on \texttt{Antmaze} and is competitive with OT on parts of \texttt{MuJoCo}, despite ignoring ordering entirely.
Third, the \texttt{Adroit} results exhibit substantially larger variance across seeds, suggesting that conclusions there are less stable and that differences among alignment choices are harder to resolve from mean scores alone.

Overall, these results are consistent with the view that much of the performance can be obtained through nearest-neighbor proximity induced by a point-to-set distance and a coarse temporal correspondence via segmentation.

\paragraph{ReBRAC Learner}
Many reward learning and IRL papers claim that reward labeling can be paired with any offline RL learners to match or surpass the learners' Oracle performance with ground truth rewards. Yet, their evaluations are often restricted to IQL \citep{ot_reward, liu2023clue, lyu2024seabo}.
We broaden the evaluation by using ReBRAC \citep{tarasov2023rebrac} as an additional downstream learner.
ReBRAC is a strong behavior-regularized actor-critic method and represents a different design choice from IQL, which helps assess whether conclusions about reward assignment depend on the downstream algorithm.

We use the ReBRAC implementation\footnote{With critic regularization being 0 and actor regularization being 1, ReBRAC can be approximately viewed as the TD3+BC algorithm \citep{td3_bc}; we report TD3+BC results in Appendix~\ref{para:td3+bc}.} from the Clean Offline RL (CORL) codebase \citep{corl_codebase}.
The overall protocol and evaluation criterion follow the IQL experiments.
In practice, ReBRAC performance is sensitive to the actor and critic regularization coefficients, so on \texttt{MuJoCo} we sweep a fixed grid of these two coefficients for each reward assignment method using an identical protocol, and report the best-performing setting within the hyperparameter grid. Details are in Appendix~\ref{para:rebrac_tuning}, where we also report \emph{untuned} default-hyperparameter results to show sensitivity.

\begin{table}[ht]
    \caption{Normalized scores with ReBRAC as the downstream RL algorithm. Hyperparameters are \textbf{tuned} according to Table~\ref{tabel:tuned_rebrac_params_mujoco} and Table~\ref{tabel:tuned_rebrac_params_antmaze}.}
    \label{tabel:rebrac_main}
    \centering
    \scalebox{0.7}
    {
    \begin{tabular}{lccccccccccccc}

\toprule
Dataset & BC & ReBRAC (oracle) & OT & TemporalOT & Seg-match & Min-Dist \\
\hline
hopper-medium-replay & 29.81 $\pm$ 2.07 & 98.1 $\pm$ 5.3 & \textbf{98.57} $\pm$ 1.6 & \textbf{97.63} $\pm$ 2.1 & 91.86 $\pm$ 2.9 & \textbf{97.43} $\pm$ 1.4 \\
hopper-medium & 53.51 $\pm$ 1.76 & 102.0 $\pm$ 1.0 & \textbf{98.63} $\pm$ 1.0 & \textbf{98.43} $\pm$ 1.2 & \textbf{96.48} $\pm$ 3.4 & \textbf{95.09} $\pm$ 7.2 \\
hopper-medium-expert & 52.30 $\pm$ 4.01 & 107.0 $\pm$ 6.4 & \textbf{109.12} $\pm$ 2.0 & \textbf{108.04} $\pm$ 3.2 & \textbf{109.18} $\pm$ 2.1 & \textbf{107.94} $\pm$ 3.3 \\
halfcheetah-medium-replay & 35.66 $\pm$ 2.33 & 51.0 $\pm$ 0.8 & \textbf{46.19} $\pm$ 0.4 & \textbf{48.30} $\pm$ 0.7 & \textbf{47.43} $\pm$ 0.9 & \textbf{48.27} $\pm$ 0.9 \\
halfcheetah-medium & 42.40 $\pm$ 0.19 & 65.6 $\pm$ 1.0 & 47.28 $\pm$ 0.3 & \textbf{50.79} $\pm$ 0.7 & 44.49 $\pm$ 1.2 & \textbf{49.87} $\pm$ 0.7 \\
halfcheetah-medium-expert & 55.95 $\pm$ 7.35 & 101.1 $\pm$ 5.2 & \textbf{92.17} $\pm$ 1.2 & \textbf{92.71} $\pm$ 1.7 & \textbf{91.87} $\pm$ 1.6 & \textbf{92.45} $\pm$ 2.6 \\
walker2d-medium-replay & 21.80 $\pm$ 10.15 & 77.3 $\pm$ 7.9 & 71.24 $\pm$ 6.1 & 72.61 $\pm$ 7.2 & \textbf{84.57} $\pm$ 10.2 & 77.04 $\pm$ 4.0 \\
walker2d-medium & 63.23 $\pm$ 16.24 & 82.5 $\pm$ 3.6 & \textbf{80.53} $\pm$ 1.1 & \textbf{81.71} $\pm$ 0.9 & \textbf{83.99} $\pm$ 0.4 & \textbf{80.77} $\pm$ 1.0 \\
walker2d-medium-expert & 98.96 $\pm$ 15.98 & 111.6 $\pm$ 0.3 & \textbf{108.72} $\pm$ 0.5 & \textbf{108.88} $\pm$ 0.5 & \textbf{108.99} $\pm$ 0.7 & \textbf{109.25} $\pm$ 0.6 \\
\bottomrule
\textbf{MuJoCo-total} & 453.6 & 796.2 & \textbf{752.46} & \textbf{759.11} & \textbf{758.85} & \textbf{758.11} \\
\bottomrule

    \end{tabular}
    }
    
\end{table}

Table~\ref{tabel:rebrac_main} reports normalized scores on the nine \texttt{MuJoCo} datasets. The ReBRAC Oracle results are taken from \citet{tarasov2023rebrac}. The main takeaway mirrors the IQL results: in the offline setting, nearest-neighbor proximity alone encodes a large portion of the learning signal required to imitate expert demonstrations. With a stronger downstream learner and well-tuned hyperparameters, the benefits from both sophisticated proximity approximations and temporal alignment largely diminish, as all four methods achieve the same performance. Moreover, Table~\ref{tabel:rebrac_main} reveals a substantial performance gap between Oracle and proximity-reward-based learning, suggesting that the performance superiority claimed in prior work may largely depend on the choice of downstream RL learner.

\subsection{Online RL Evaluation}
\label{sec:exps_online_rl}
We next evaluate our proposed reward relabeling methods in the online reinforcement learning setting. As in \citet{temporal_ot}, we assume that: $(i)$ two expert trajectories are available; $(ii)$ no non-expert trajectories are available; $(iii)$ the agent interacts with the environment during training to collect reward-less state and action trajectory data. This setting tests whether reward assignment methods can provide a usable learning signal for long-horizon exploration and policy optimization when the replay buffer is populated on the fly. Experiments are conducted on the MetaWorld benchmark \citep{metaworld2020}.

For policy learning, we adopt DrQ-v2 \citep{drqv2}, an off-policy actor-critic algorithm that combines sample-efficient Q-learning with strong visual data augmentation. Policies are trained from scratch using only the rewards generated by the four relabeling methods, without access to the environment’s ground-truth rewards. Note that, unlike offline RL evaluations, where we can compute global dataset statistics to normalize the relabeled rewards, online RL maintains a dynamic replay buffer and thus uses a different strategy for reward post-processing. More details about task configurations, implementation details, and training hyperparameters can be found in Appendix~\ref{appx:metaworld_setup}.

In the MetaWorld benchmark, all task episodes have a fixed length, which reduces the Segment-matching method to one-to-one matching between identical agent and expert timesteps. While this preserves temporal order, it is less aware of the local trajectory structure and risks large reward variance along an episode. A simple remedy is to smooth the relabeled episodic rewards with a Gaussian kernel. We therefore report two variants: Seg-Match (raw), which applies the segment correspondence directly, and Seg-Match (smooth), which additionally applies the Gaussian kernel. This isolates whether the online gains come from temporal correspondence itself or merely from reward smoothing.

In Table~\ref{tabel:metaworld_main}, we report the success rate multiplied by 100 for each task, averaged over 5 random seeds. As in the offline experiments, we report the average over the last four evaluations. The DrQ-v2 Oracle results are taken from \citet{temporal_ot} with ground-truth sparse task reward. We omit the Min-Dist reward because it yields a \emph{zero} success rate across all tasks. 



\begin{table}[ht]
    \caption{Success rates (\%) for online RL on MetaWorld with DrQ-v2 as the downstream RL algorithm. DrQ-v2 (Oracle) denotes performance obtained from training with ground-truth \textit{sparse} task reward.}
    \label{tabel:metaworld_main}
    \centering
    \scalebox{0.8}{
    \begin{tabular}{lccccc}
    
\toprule
Dataset & DrQ-v2 (Oracle) & OT & TemporalOT & Seg-Match (raw) & Seg-Match (smooth) \\
\hline
window-open-v2 & 85.6 $\pm$ 12.2 & 15.75 $\pm$ 8.5 & 21.25 $\pm$ 5.4 & \textbf{22.25} $\pm$ 4.1 & \textbf{22.75} $\pm$ 4.7 \\
door-open-v2 & 0.0 $\pm$ 0.0  & 73.05 $\pm$ 12.2 & 75.20 $\pm$ 13.8 & \textbf{79.90} $\pm$ 6.4 & \textbf{79.35} $\pm$ 3.6 \\
door-lock-v2 & 86.2 $\pm$ 12.4 & 43.80 $\pm$ 36.0 & \textbf{67.10} $\pm$ 7.1 & 45.95 $\pm$ 16.4 & 62.30 $\pm$ 5.9 \\
push-v2 & 1.0 $\pm$ 0.7 & 12.55 $\pm$ 3.5 & \textbf{17.12} $\pm$ 2.3 & 14.50 $\pm$ 3.0 & 15.45 $\pm$ 3.6 \\
stick-push-v2 & 0.0 $\pm$ 0.0 & \textbf{96.95} $\pm$ 2.1 & \textbf{98.19} $\pm$ 1.7 & \textbf{98.05} $\pm$ 1.1 & \textbf{97.80} $\pm$ 1.3 \\
basketball-v3 & 0.0 $\pm$ 0.0 & 73.90 $\pm$ 35.6 & \textbf{94.70} $\pm$ 2.5 & 74.95 $\pm$ 13.6 & 85.85 $\pm$ 4.6 \\
button-press-topdown-v2 & 14.0 $\pm$ 18.5 & \textbf{65.10} $\pm$ 6.2 & \textbf{64.00} $\pm$ 6.6 & \textbf{65.50} $\pm$ 5.0 & \textbf{65.65} $\pm$ 5.1 \\
hand-insert-v2 & 0.8 $\pm$ 1.6 & 8.30 $\pm$ 2.8 & \textbf{9.25} $\pm$ 2.9 & \textbf{8.90} $\pm$ 2.5 & \textbf{8.85} $\pm$ 3.0 \\
lever-pull-v2 & 0.0 $\pm$ 0.0 & 2.50 $\pm$ 1.4 & 2.80 $\pm$ 1.2 & \textbf{3.40} $\pm$ 2.1 & 1.50 $\pm$ 1.3 \\
\bottomrule
\textbf{Metaworld-total} & 187.6 & 391.90 & \textbf{449.61} & 413.40 & \textbf{439.50} \\
\bottomrule
    \end{tabular}
    }
\end{table}

The results show that both Seg-Match variants outperform OT and Min-Dist across nearly every task. Min-Dist fails completely, while raw Seg-Match already exceeds OT on all nine tasks, indicating that temporal correspondence is what makes the proximity signal usable at all. Smoothing then reduces variance on the highest-variance tasks (e.g., door-lock and basketball), lifting the total to a level competitive with TemporalOT.

This performance gap also suggests that in long-horizon manipulation tasks, the main failure mode is not an inaccurate proximity approximation, but an underspecified notion of correspondence: without temporal structure, relabeling can assign spuriously high credit to states that are close to some expert state yet lie on an incompatible phase of the task (e.g., initial states), producing a noisy learning signal and weaker exploration. Enforcing a temporal correspondence thus stabilizes the proximity signal, so that the agent is encouraged to make progress along the demonstrated behavior rather than being rewarded for matching arbitrary expert states out of context.

Overall, these online results support the broader conclusion that reward assignment benefits primarily from designing the correspondence structure, while the additional complexity of solving for a global transport plan is not always the determining factor.

\subsection{Offline Multi-Expert Scaling}
\label{sec:multi-expert-main}

\begin{figure}[ht]
    \begin{center}
       \includegraphics[width=0.9\linewidth]{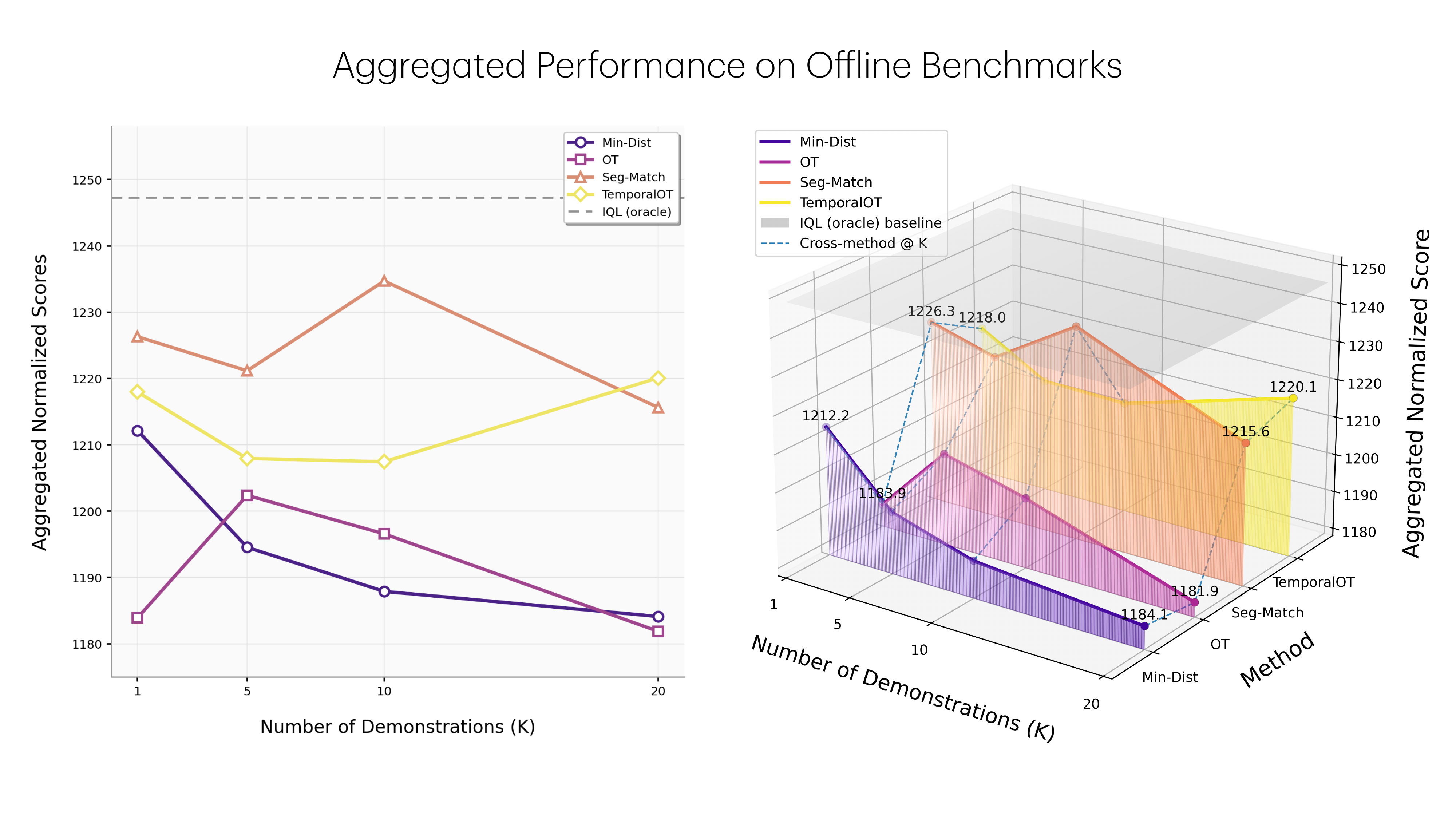}
    \end{center}
    \caption{Aggregated D4RL performance as a function of the number of experts ($K$).} 
    \label{fig:k-aggregated-main}
\end{figure}

We study how performance scales with the number of expert demonstrations $K \in \{1,5,10,20\}$ in the offline setting, using IQL as the downstream learner.
For $K>1$, following \citet{ot_reward, temporal_ot}, we compute proxy rewards with respect to \emph{each} expert trajectory and retain the set of relabeled rewards from the expert that yields the highest relabeled episode return. Figure~\ref{fig:k-aggregated-main} reports normalized scores aggregated across all offline benchmarks. We run 5 seeds for these experiments. 

Figure~\ref{fig:k-aggregated-main} highlights three observations: \textbf{(i)} Temporal correspondence improves performance across all $K$ values. The curves for the Seg-Match and TemporalOT lie consistently above those for the Min-Dist reward and OT reward. \textbf{(ii)} Temporal alignment maintains performance with increasing $K$. Methods lacking the temporal structure, Min-Dist and OT, exhibit declining performance as $K$ increases. \textbf{(iii)} Stronger proximity approximation is not the dominant factor in this scaling regime. The Seg-Match reward remains competitive across most $K$, despite avoiding transport optimization. 

Multi-expert settings expose a failure mode in the absence of temporal alignment. When multiple demonstrations are available, temporally invariant proximity approximation can yield close distances corresponding to inconsistent phases across demonstrations, which degrades the learning signal as $K$ grows. This scaling study suggests that temporal correspondence can matter more than how proximity is approximated. In Appendix~\ref{appx: multi-demo}, we provide the same scaling plots separately for each offline domain.

\subsection{Why Can Simple Proximity Be Sufficient?}
\label{sec:thm}
Inspired by \citet{yu2022uds}, this section provides a bound in Theorem~\ref{thm:bound} that helps interpret when rewards labeled by a point-to-set distance may encode enough information required to achieve similar performance as OT rewards in the offline setting. The full theorem statement and proof are provided in Appendix~\ref{appx:theory}.

\begin{theorem}[Sufficiency Bound; condensed version] \label{thm:bound}
Under mild regularity conditions on concentration of relabeled OT rewards $r_{s,a}^{ot}$, with high probability $\ge 1-\delta$, the optimal trajectory-matching policy learned by OT performs no better than Min-Dist rewards by the following bound:
\begin{align*}
J(\pi_{ot}^\ast) - J(\pi_{min}^\ast) &\leq \frac{1}{1-\gamma} \mathbb{E}_{d_{\D}^{\pi_\beta}}\bigl[\rho(s,a) \cdot \left(\mathcal{R}(s,a)-r_{s,a}^{min}\right)\bigr] \\
& - \frac{1}{1-\gamma}\mathbb{E}_{d_{\D}^{\pi_{min}^\ast}}\bigl[\rho(s,a) \cdot \left(\mathcal{R}(s,a)-r_{s,a}^{min}\right)\bigr] \\
& - \frac{\alpha}{1-\gamma}\Div(\pi_{min}^\ast, \pi_\beta) + \text{sampling error}
\end{align*}

where $J(\pi)$ is the expected discounted return of $\pi$ in MDP; and $d_{\D}^{\pi}(s,a)$ is the marginal state-action distribution of $\pi$ induced by dataset $\D$; $\rho\in[0,1]$ estimates the percentage of mismatch between OT and Min-Dist relabeled rewards in the dataset; $\mathcal{R}$ is the average stationary OT reward function; $\Div$ is a divergence metric.
\end{theorem}

Intuitively, $J(\pi)$ can be interpreted as the average proximity to the expert trajectory in Wasserstein space when following $\pi$. The larger $J(\pi)$ is, the more closely $\pi$ follows the expert trajectory. Thus, Theorem~\ref{thm:bound} implies that the optimal policy learned by Min-Dist rewards is sufficient to encode the trajectory-matching ability, resulting in similar performance as the optimal OT policy when the bound is small. We comment on several scenarios where this bound may become small:

\begin{itemize}[left=0pt]
    \item \textbf{Low-quality data regime}: Given a low-quality dataset, the marginal state-action distribution of the behavior policy $\pi_\beta$ concentrates around low-quality transitions whose proximity rewards $\mathcal{R}(s,a)$ are small. Thus, the first term of the bound becomes small. The second term is also likely to be small, as $d^{\pi_{min}^\ast}$ concentrates around expert transitions and thus their proximity rewards are high. Therefore, the bound becomes small. Indeed, Table~\ref{tabel:iql_main} shows that OT achieves 160.99 on three \texttt{MuJoCo-medium-replay} datasets, while Min-Dist achieves 200.16.
    \item \textbf{High-quality data regime}: In this case, both $d^{\pi_\beta}$ and $d^{\pi_{min}^\ast}$ should concentrate around the expert transitions. For near-expert states, it is very likely that the reward difference between $\mathcal{R}(s,a)$ and $r_{s,a}^{min}$ is close to zero. Therefore, the bound should be small in this case. In Table~\ref{tabel:iql_main}, OT (73.43) and Min-Dist (71.47) perform similarly on \texttt{Adroit-human} datasets.
    \item \textbf{Better offline RL Learner}: A better algorithm can learn a policy from Min-Dist rewards that approximates expert behavior better. In this case, the third term of the bound becomes small. Considering the ReBRAC results in Table~\ref{tabel:rebrac_main}, Min-Dist (758.11) indeed exhibits similar performance to OT (752.46).
\end{itemize}

\section{Related Work} \label{sec: related_work}
Inverse reinforcement learning (IRL) seeks to infer reward functions from expert demonstrations under which the expert behavior is optimal \citep{Ng2000irl_algos, arora2020survey_irl}. Several approaches instantiate this idea with different reward-learning mechanisms. SQIL reduces online imitation to Q-learning by assigning reward 1 to the expert and reward 0 to online trajectories \citep{reddy2020sqil}, while ORIL learns rewards with a discriminator and then applies offline RL \citep{oril}. Similarly, UDS \citep{yu2022uds} assigns zero reward to all offline unlabeled trajectories while assuming access to another dataset with ground-truth rewards. SMODICE matches state-occupancy measures via an $f$-divergence objective \citep{ma2022smodice}, and IQ-Learn learns a single Q-function that implicitly represents both reward and policy, avoiding adversarial reward–policy optimization \citep{garg2021iq_learn}. Representation-based methods such as CLUE learn latent spaces and define proximity rewards in that space \citep{liu2023clue}. While effective, these methods typically require additional training objectives, representation learning, or sufficient demonstrations\footnote{We report additional comparisons to related methods from the literature in Appendix~\ref{appx: literature}.}.

Another line of work constructs proxy rewards directly from distances between learner and expert states. \citet{lyu2024seabo} drops learned components entirely, assigning rewards by searching for the nearest expert state-action concatenations. Optimal transport (OT) aligns trajectories through a minimum-cost coupling defined by the Wasserstein distance. \citet{xiao2019wail} connects adversarial imitation learning with OT via its dual formulation, while \citet{dadashi2021pwil} minimizes the Wasserstein distance through its primal formulation. However, OT-based rewards \citep{ot_reward} ignore temporal ordering, motivating methods that introduce temporal structure. In addition to TemporalOT \citep{temporal_ot}, ORCA assigns rewards based on ordered coverage of temporally misaligned video demonstration states rather than frame-level proximity \citep{huey2025orca}.

OT-based trajectory matching has also been applied to robotic policy fine-tuning \citep{haldar2022watch_match, haldar2023teach_fish}, cross-domain imitation learning \citep{fickinger2022cross_domain_ot, lyu2025crossdomain_ot}, and formulations of offline RL as an optimal transport problem \citep{asadulaev2024rethinking_ot}. These works highlight the flexibility of distance-based alignment while underscoring the importance of how proximity and temporal structure are incorporated into reward assignment.

\section{Conclusion}
This paper studies proximity-based reward assignment from demonstrations through the question: what minimal structure must a reward rule encode to enable effective RL? We analyze this problem along two design choices: how proximity to demonstrations is approximated and whether temporal correspondence is enforced. Across 32 benchmarks spanning offline and online settings and multiple downstream learners, the results convey \textbf{two practical takeaways:} (1) nearest-neighbor proximity to a demonstration alone often provides a strong learning signal in offline regimes, while (2) lightweight temporal correspondence becomes essential when multiple demonstrations are present and vary in phase, or online exploration makes credit assignment fragile and dependent on task progress.  

More broadly, our motivation is that solving a reward assignment problem should begin with understanding the structure of the problem itself. In particular, one should ask which ingredients are actually needed to produce an effective learning signal, rather than defaulting to more complex machinery. Methods based on global coupling or transport-plan optimization are powerful and broadly applicable, but their use is not always guided by a clear understanding of which structural properties are truly necessary under limited demonstrations. The two design choices studied in this paper are therefore not intended as a complete reward assignment algorithm. Rather, they serve as simple diagnostic axes that reveal which ingredients matter, when they matter, and why. We hope these findings encourage more careful thinking about the structure of reward assignment problems, and suggest that simple reward rules can serve as a strong starting point before more complex methods are introduced. One limitation of our study is that proximity is measured directly in the original state space \citep{liu2023clue, yang2026latent_wail}. A natural extension is therefore to consider a third orthogonal design dimension, representation learning, and investigate the minimal latent properties required for representations to support effective reward assignment. We hope this work lays a foundation for future research on reward assignment beyond state-space matching, including how reward design can shape learning in representation space itself.

\subsubsection*{Acknowledgments}
This work was supported in part by NYU Abu Dhabi Center for Artificial Intelligence and Robotics and Center for Interdisciplinary Data Science and AI, funded by Tamkeen under the Research Institute Award CG010. It was also partially supported by Shanghai Frontiers Science Center of Artificial Intelligence and Deep Learning at NYU Shanghai. Computational resources were provided in part by NYU IT High-Performance Computing resources and services.

\bibliography{rlj}
\bibliographystyle{rlj}

\newpage

\beginSupplementaryMaterials
\appendix
\section{Offline RL Evaluations}
\subsection{Reward Post-processing}
\label{appx:offline_postprocessing}

To improve training stability for value-based offline RL, we apply the same post-processing structure used in prior OT reward labeling pipelines \citep{dadashi2021pwil, ot_reward}. Post-processing is applied to the raw proxy reward produced by each reward assignment method and consists of two stages: exponential squashing followed by an affine rescaling.

\paragraph{Exponential squashing.}
For any raw reward $r_t$, we apply
\[
r_t \leftarrow \alpha \cdot \exp(\beta \cdot r_t),
\]
where $\alpha>0$ sets the overall scale and $\beta>0$ controls sharpness. For OT-based approaches, we used the best-performing parameters $(\alpha, \beta) = (5,5)$. Appendix A.6 of \citet{ot_reward} shows that OT reward with $(\alpha, \beta) = (1, 1)$ results in a clear performance drop across \emph{all} locomotion datasets. For Min-Dist and Seg-Match, we verified that (1,1) and (5,5) perform similarly, so we kept the primitive choice (1,1). Thus, each method is evaluated under the hyperparameters known to represent it well, rather than eclipsing OT with an underperforming config.

\paragraph{Dataset-level rescaling and bias.}
After squashing, we apply an affine transformation
\[
r_t \leftarrow \texttt{reward\_scale} \cdot r_t + \texttt{reward\_bias},
\]
as commonly done in offline RL \citep{iql, td3_bc, ot_reward}. The scale is computed from relabeled episodic returns over the offline dataset:
\[
\texttt{reward\_scale} = \frac{1000}{\texttt{max\_return} - \texttt{min\_return}},
\]
where $\texttt{max\_return}$ and $\texttt{min\_return}$ are the maximum and minimum relabeled returns across trajectories in the dataset. We adopt the same fixed bias as \citet{ot_reward}: $\texttt{reward\_bias}=0$ for \texttt{MuJoCo} and \texttt{Adroit}, and $\texttt{reward\_bias}=-2$ for \texttt{Antmaze}.

These two post-processing stages are applied to all compared methods; the only method-dependent choice in this appendix is the squashing parameters, which follow published OT defaults for OT/TemporalOT and a single untuned default for Seg-Match/Min-Dist.
\subsection{IQL Experiments}
\label{appx:hyperparam_offline}

We use the official implementation of OT reward \citep{ot_reward} and thus adopt all the hyperparameters. Table \ref{tabel:iql_hyperparameters} summarizes them and also includes the information about reward labeling and computation resources for each individual experiment. 

\begin{table}[ht]
    \caption{Hyperparameters for IQL offline evaluations.}
    \label{tabel:iql_hyperparameters}
    \centering
    \scalebox{0.75}
    {
    \begin{tabular}{cllccccccccccc}

\toprule
    & \textbf{Hyperparameter} & \textbf{Value} \\
\hline
    & Total training steps & 1e6 \\
    & Evaluation frequency & 1e4, MuJoCo, Adroit \\
    &   & 5e4, Antmaze \\
 Training   & Evaluation episodes & 10, MuJoCo, Adroit \\
    &   & 100, Antmaze \\
    & Batch size & 256 \\
    & Seeds & \texttt{range(10)} \\
\hline
    & Hidden layers & (256, 256) \\
Network Architecture    & Dropout & None \\
    & Network initialization & orthogonal \\
\hline
    & Discount factor & 0.99 \\
    & Optimizer & Adam \\
    & Policy learning rate & $3 \times 10^{-4}$, cosine decay to 0 \\
    & Critic learning rate & $3 \times 10^{-4}$ \\
    & Value learning rate & $3 \times 10^{-4}$ \\
    & Target network update rate & $5 \times 10^{-3}$ \\
 IQL   & Temperature & 3.0, MuJoCo \\
    &   & 0.5, Adroit \\
    &   & 10, Antmaze \\
    & Expectile & 0.7, MuJoCo, Adroit \\
    &   & 0.9, Antmaze \\
    & State normalization & True, Mujoco, Adroit\\
    &   & False, Antmaze \\
\hline
    & Episode length $T$ & 1000 \\
    & Cost function & cosine distance \\
Reward Labeling    & Squashing function & $r \leftarrow \alpha \exp(\beta r)$, OT \& TemporalOT use $\alpha=5,\ \beta=5$ \\
&   & $r \leftarrow \alpha \exp(\beta r)$, Seg-Match \& Min-Dist use $\alpha=1,\ \beta=1$ \\
    & Number of experts & 1 \\
\hline
    & Compute resources & Intel(R) Xeon(R) Platinum 8268 CPU \\
Computation    & Number of CPU workers & 1 \\
    & Requested compute memory & 8 GB \\
    & Approximate average execution time & 6 hrs \\
\bottomrule

    \end{tabular}
    }
    
\end{table}

\paragraph{Evaluations with Euclidean Distance}
\label{para:iql_eulidean}
Throughout the main body, we adopt the commonly used cosine distance as the similarity metric for any two states when IRL methods compute the pseudo-rewards. Here, we also include evaluations by using Euclidean distance for IRL reward labeling in Table \ref{tabel:iql_euclidean}. Note that compared with using cosine distance, the performance in general declines, which indicates that Euclidean distance may struggle to embed and capture the geometric property, e.g., proximity or similarity, among states. Since none of the methods work for \texttt{Adroit}'s Door, Hammer, and Relocate tasks, we omit those evaluations. 

\begin{table}[ht]
    \caption{Normalized scores with IQL as downstream RL algorithm. \textbf{Euclidean distance} is used instead of cosine distance.}
    \label{tabel:iql_euclidean}
    \centering
    \scalebox{0.75}
    {
    \begin{tabular}{lccccccccccccc}

\toprule
Dataset & IQL (oracle) & OT & TemporalOT & Seg-match & Min-Dist \\
\hline
hopper-medium-replay & 94.66 $\pm$ 8.2 & \textbf{95.03} $\pm$ 1.3 & \textbf{96.05} $\pm$ 1.0 & \textbf{97.10} $\pm$ 1.0 & \textbf{96.17} $\pm$ 1.3 \\
hopper-medium & 63.50 $\pm$ 4.8 & \textbf{75.16} $\pm$ 4.5 & \textbf{73.78} $\pm$ 3.8 & \textbf{77.52} $\pm$ 5.0 & 71.59 $\pm$ 3.7 \\
hopper-medium-expert & 103.67 $\pm$ 8.9 & \textbf{108.12} $\pm$ 3.2 & \textbf{107.58} $\pm$ 4.6 & \textbf{109.14} $\pm$ 2.7 & 103.09 $\pm$ 20.1 \\
halfcheetah-medium-replay & 43.12 $\pm$ 1.6 & 39.01 $\pm$ 2.0 & 37.07 $\pm$ 2.4 & 38.46 $\pm$ 2.6 & \textbf{42.75} $\pm$ 1.1 \\
halfcheetah-medium & 47.49 $\pm$ 0.3 & \textbf{43.28} $\pm$ 0.7 & 42.56 $\pm$ 0.5 & 42.76 $\pm$ 0.3 & \textbf{45.11} $\pm$ 0.6 \\
halfcheetah-medium-expert & 91.24 $\pm$ 2.3 & \textbf{92.08} $\pm$ 1.3 & 58.51 $\pm$ 6.7 & \textbf{91.44} $\pm$ 1.5 & \textbf{92.56} $\pm$ 1.4 \\
walker2d-medium-replay & 73.79 $\pm$ 8.9 & \textbf{55.09} $\pm$ 22.0 & \textbf{56.72} $\pm$ 14.7 & 52.16 $\pm$ 16.2 & 49.97 $\pm$ 24.4 \\
walker2d-medium & 80.57 $\pm$ 4.2 & \textbf{79.65} $\pm$ 0.7 & \textbf{77.78} $\pm$ 3.4 & 72.50 $\pm$ 4.8 & \textbf{80.17} $\pm$ 0.5 \\
walker2d-medium-expert & 111.06 $\pm$ 0.2 & \textbf{110.76} $\pm$ 0.1 & \textbf{110.14} $\pm$ 0.1 & \textbf{109.47} $\pm$ 0.2 & \textbf{110.85} $\pm$ 0.2 \\
\bottomrule
\textbf{MuJoCo-total} & 709.10 & \textbf{698.18} & 660.19 & \textbf{690.54} & \textbf{692.27} \\
\bottomrule
\toprule
pen-human & 69.86 $\pm$ 17.2 & \textbf{67.69} $\pm$ 17.0 & \textbf{69.67} $\pm$ 22.5 & \textbf{69.60} $\pm$ 19.3 & \textbf{69.40} $\pm$ 21.1 \\
pen-cloned & 70.96 $\pm$ 20.3 & 58.31 $\pm$ 21.4 & 57.44 $\pm$ 18.3 & \textbf{63.71} $\pm$ 19.4 & \textbf{61.02} $\pm$ 20.0 \\
\bottomrule
\textbf{Adroit-total} & 140.82 & 126.00 & \textbf{127.12} & \textbf{133.32} & \textbf{130.42} \\
\bottomrule
\toprule
antmaze-umaze-play & 86.35 $\pm$ 3.5 & 89.03 $\pm$ 2.9 & \textbf{90.48} $\pm$ 2.6 & \textbf{91.80} $\pm$ 2.9 & \textbf{93.83} $\pm$ 1.6 \\
antmaze-umaze-diverse & 64.05 $\pm$ 8.1 & 62.95 $\pm$ 12.2 & 63.92 $\pm$ 8.3 & \textbf{69.00} $\pm$ 7.6 & \textbf{70.75} $\pm$ 5.7 \\
antmaze-medium-play & 75.45 $\pm$ 4.7 & 67.55 $\pm$ 5.1 & \textbf{70.72} $\pm$ 6.2 & \textbf{73.28} $\pm$ 4.0 & 63.03 $\pm$ 5.4 \\
antmaze-medium-diverse & 72.92 $\pm$ 5.4 & \textbf{67.17} $\pm$ 5.1 & \textbf{69.72} $\pm$ 4.8 & \textbf{69.12} $\pm$ 4.8 & \textbf{68.15} $\pm$ 4.7 \\
antmaze-large-play & 46.38 $\pm$ 5.5 & \textbf{49.17} $\pm$ 5.7 & \textbf{48.38} $\pm$ 6.3 & 42.90 $\pm$ 6.3 & 46.63 $\pm$ 4.7 \\
antmaze-large-diverse & 45.15 $\pm$ 5.4 & \textbf{50.97} $\pm$ 4.9 & 48.23 $\pm$ 5.9 & 45.75 $\pm$ 5.7 & 45.55 $\pm$ 6.3 \\
\bottomrule
\textbf{Antmaze-total} & 390.30 & \textbf{386.85} & \textbf{391.45} & \textbf{391.85} & \textbf{387.93} \\
\bottomrule

    \end{tabular}
    }
    
\end{table}

\subsection{ReBRAC Experiments}
\paragraph{Hyperparameter Settings} Because ReBRAC tunes hyperparameters for each individual dataset in D4RL, we refer readers to the official ReBRAC implementation\footnote{\url{https://github.com/tinkoff-ai/ReBRAC/tree/public-release/configs/rebrac}} to learn all the details. We keep the default ReBRAC hyperparameters unchanged for the \texttt{MuJoCo} experiments in Table \ref{tabel:rebrac_mujoco_untuned}, but tune the BC regularization of both critic and actor for experiments in Table~\ref{tabel:rebrac_main} and Table~\ref{tabel:rebrac_tuned}. The setup for ReBRAC reward labeling and computation requirements are the same as IQL experiments in Table \ref{tabel:iql_hyperparameters}.

\paragraph{Hyperparameter Tuning for ReBRAC}
\label{para:rebrac_tuning}
ReBRAC substantially tuned hyperparameters. e.g., critic BC regularization strength, actor BC regularization strength, critic/actor learning rate, and batch size. As a result, each D4RL dataset inherits a set of optimal hyperparameters. However, such tuned hyperparameters are only optimal when training ReBRAC with ground-truth task rewards provided by the datasets. For example, D4RL \texttt{Antmaze} domain provides a sparse reward signal of 1 when the ant agent successfully reaches the goal before the environment timeout; otherwise, 0. To overcome the issue of reward sparsity, ReBRAC uses a larger discount factor $\gamma=0.999$ to propagate the reward signal through TD learning faster. However, such a large discount factor is not suitable for dense reward tasks, like \texttt{MuJoCo}. So \texttt{MuJoCo} and \texttt{Adroit} use $\gamma=0.99$. For the four IRL rewards discussed throughout this paper, they assign dense rewards to task episodes, including \texttt{Antmaze} domain. Therefore, $\gamma=0.999$ is incompatible for the four reward labeling methods and indeed all of them fail in this case. In addition to the influence of the discount factor, performance is also sensitive to the BC regularizer strengths for both critic and actor training. This explains the abrupt failure of Segment-matching reward on \texttt{halfcheetah-medium} dataset in Table~\ref{tabel:rebrac_mujoco_untuned} if we adopt the default untuned regularizations. Despite this performance sensitivity, Min-Dist remaining strongest here further supports our main offline claim that proximity alone already provides a strong reward signal.

\begin{table}[h]
    \caption{Normalized scores with ReBRAC as the downstream RL algorithm. The default \textbf{untuned} ReBRAC hyperparameters are used in this table.}
    \label{tabel:rebrac_mujoco_untuned}
    \centering
    \scalebox{0.7}
    {
    \begin{tabular}{lccccccccccccc}

\toprule
Dataset & BC & ReBRAC (oracle) & OT & TemporalOT & Seg-match & Min-Dist \\
\hline
hopper-medium-replay & 29.81 $\pm$ 2.07 & 98.1 $\pm$ 5.3 & \textbf{95.60} $\pm$ 2.4 & \textbf{96.35} $\pm$ 2.5 & 90.38 $\pm$ 4.0 & \textbf{95.99} $\pm$ 2.3 \\
hopper-medium & 53.51 $\pm$ 1.76 & 102.0 $\pm$ 1.0 & \textbf{95.00} $\pm$ 6.8 & \textbf{96.01} $\pm$ 4.8 & 73.68 $\pm$ 30.7 & \textbf{97.36} $\pm$ 2.8 \\
hopper-medium-expert & 52.30 $\pm$ 4.01 & 107.0 $\pm$ 6.4 & \textbf{104.75} $\pm$ 6.2 & 99.48 $\pm$ 6.5 & \textbf{108.17} $\pm$ 2.2 & \textbf{109.33} $\pm$ 1.2 \\
halfcheetah-medium-replay & 35.66 $\pm$ 2.33 & 51.0 $\pm$ 0.8 & 40.06 $\pm$ 1.3 & 42.67 $\pm$ 0.9 & \textbf{45.42} $\pm$ 1.5 & \textbf{43.54} $\pm$ 0.4 \\
halfcheetah-medium & 42.40 $\pm$ 0.19 & 65.6 $\pm$ 1.0 & 47.01 $\pm$ 0.3 & \textbf{49.69} $\pm$ 0.5 & 2.36 $\pm$ 1.1 & \textbf{48.92} $\pm$ 0.5 \\
halfcheetah-medium-expert & 55.95 $\pm$ 7.35 & 101.1 $\pm$ 5.2 & \textbf{92.27} $\pm$ 1.1 & \textbf{92.78} $\pm$ 1.6 & \textbf{89.99} $\pm$ 2.6 & \textbf{92.66} $\pm$ 1.7 \\
walker2d-medium-replay & 21.80 $\pm$ 10.15 & 77.3 $\pm$ 7.9 & 68.25 $\pm$ 5.7 & 67.80 $\pm$ 6.8 & \textbf{80.96} $\pm$ 12.8 & 72.36 $\pm$ 5.5 \\
walker2d-medium & 63.23 $\pm$ 16.24 & 82.5 $\pm$ 3.6 & \textbf{78.64} $\pm$ 1.5 & \textbf{79.08} $\pm$ 1.5 & \textbf{80.45} $\pm$ 1.4 & \textbf{78.96} $\pm$ 1.1 \\
walker2d-medium-expert & 98.96 $\pm$ 15.98 & 111.6 $\pm$ 0.3 & \textbf{107.05} $\pm$ 2.3 & \textbf{107.57} $\pm$ 2.2 & \textbf{108.99} $\pm$ 0.7 & \textbf{109.23} $\pm$ 0.4 \\
\bottomrule
\textbf{MuJoCo-total} & 453.6 & 796.2 & \textbf{728.63} & \textbf{731.43} & 680.40 & \textbf{748.37} \\
\bottomrule

    \end{tabular}
    }
    
\end{table}

Next, we wonder what the performance looks like if we tune the hyperparameters for each reward function and for each dataset:
\begin{itemize}
    \item \texttt{MuJoCo}: We use a batch size of 256. And we search for the best actor BC-regularization coefficient in $\{0.001, 0.01, 0.05, 0.1, 0.5\}$, and search for the best critic BC-regularization coefficient in $\{0, 0.001, 0.01, 0.1, 0.5\}$, for each MuJoCo dataset.
    \item \texttt{Adroit-pen} \& \texttt{pen-cloned}: We search for the best actor BC-regularization coefficient in $\{0.001, 0.05, 0.01, 0.1\}$, and search for the best critic BC-regularization coefficient in $\{0, 0.001, 0.01, 0.1, 0.5\}$.
    \item \texttt{Antmaze}: We use a discount factor $\gamma=0.99$, and set the learning rate of both actor and critic to be $3e^{-4}$. We then search for the best actor BC-regularization coefficient in $\{0.001, 0.002, 0.01, 0.02, 0.1, 0.2\}$, and search for the best critic BC-regularization coefficient in $\{0, 0.001, 0.01, 0.1\}$, for each antmaze dataset.
\end{itemize}

The chosen hyperparameters are reported in Table~\ref{tabel:tuned_rebrac_params_mujoco} and Table~\ref{tabel:tuned_rebrac_params_antmaze}. Table~\ref{tabel:rebrac_main} and \ref{tabel:rebrac_tuned} summarize the tuned results for OT reward and Segment-matching reward. We also include the oracle ReBRAC results for comparison. Instead of 10 seeds for experiments in the main body, we run 3 seeds for each experiment in Table~\ref{tabel:rebrac_tuned}. After hyperparameter tuning, we can see that the overall performance on Antmaze is much better than the untuned results for both OT and Segment-matching. Eventually, their total scores match each other. However, we still observe a large gap the comparing them with the oracle results. On the one hand, this may suffer from a large variance when running on only 3 seeds. On the other hand, a more sophisticated IRL method than both OT and Segment-matching may be required to close this gap in future work.

\begin{table}[ht]
    \caption{Normalized scores with ReBRAC as downstream RL algorithm. ReBRAC hyperparameters are tuned for datasets presented in this table, as well as for both OT reward and Segment-matching reward.}
    \label{tabel:rebrac_tuned}
    \centering
    \scalebox{0.75}
    {
    \begin{tabular}{lccccccccccccc}
\toprule
Dataset & ReBRAC (oracle) & OT & Seg-match \\
\hline
pen-human & 103.5 $\pm$ 14.1 & \textbf{83.03} $\pm$ 6.56 & 70.25 $\pm$ 14.71 \\
pen-cloned & 91.8 $\pm$ 21.7 & \textbf{69.11} $\pm$ 15.02 & 46.57 $\pm$ 7.60 \\
antmaze-umaze-play & 97.8 $\pm$ 1.0 & \textbf{94.25} $\pm$ 3.18 & \textbf{93.42} $\pm$ 2.78 \\
antmaze-umaze-diverse & 88.3 $\pm$ 13.0 & 49.83 $\pm$ 5.8 & \textbf{62.75} $\pm$ 6.4 \\
antmaze-medium-play & 84.0 $\pm$ 4.2 & 59.17 $\pm$ 7.3 & \textbf{81.00} $\pm$ 3.2 \\
antmaze-medium-diverse & 76.3 $\pm$ 13.5 & 33.17 $\pm$ 25.6 & \textbf{43.08} $\pm$ 12.5 \\
antmaze-large-play & 60.4 $\pm$ 26.1 & \textbf{61.58} $\pm$ 6.1 & 42.75 $\pm$ 31.4 \\
antmaze-large-diverse & 54.4 $\pm$ 25.1 & \textbf{32.33} $\pm$ 14.6 & \textbf{33.25} $\pm$ 24.0 \\
\bottomrule
\textbf{Tuned-total} & 656.5 & \textbf{482.47} & \textbf{473.07} \\
\bottomrule

    \end{tabular}
    }
    
\end{table}

\begin{table}[h]
    \caption{Normalized scores with TD3+BC as downstream RL algorithm.}
    \label{tabel:td3_bc_appx}
    \centering
    \scalebox{0.7}
    {
    \begin{tabular}{lccccccccccccc}

\toprule
Dataset & TD3+BC (oracle) & OT & TemporalOT & Seg-match & Min-Dist \\
\hline
hopper-medium-replay & 74.94 $\pm$ 22.6 & \textbf{89.63} $\pm$ 6.4 & \textbf{92.11} $\pm$ 6.0 & 84.02 $\pm$ 18.6 & 34.65 $\pm$ 10.7 \\
hopper-medium & 67.04 $\pm$ 7.4 & 83.56 $\pm$ 4.9 & 83.84 $\pm$ 5.7 & \textbf{91.05} $\pm$ 4.6 & \textbf{87.06} $\pm$ 4.6 \\
hopper-medium-expert & 104.62 $\pm$ 7.8 & 102.30 $\pm$ 3.9 & 100.94 $\pm$ 6.4 & \textbf{107.83} $\pm$ 8.3 & \textbf{109.69} $\pm$ 3.8 \\
halfcheetah-medium-replay & 46.47 $\pm$ 0.6 & 39.20 $\pm$ 1.4 & \textbf{40.76} $\pm$ 0.6 & \textbf{40.73} $\pm$ 2.6 & \textbf{41.28} $\pm$ 0.6 \\
halfcheetah-medium & 50.97 $\pm$ 0.4 & \textbf{42.89} $\pm$ 0.4 & \textbf{44.18} $\pm$ 0.3 & \textbf{45.00} $\pm$ 0.3 & \textbf{44.90} $\pm$ 0.3 \\
halfcheetah-medium-expert & 81.69 $\pm$ 6.9 & 86.50 $\pm$ 2.8 & \textbf{91.95} $\pm$ 1.8 & 86.21 $\pm$ 3.8 & \textbf{92.74} $\pm$ 1.7 \\
walker2d-medium-replay & 85.00 $\pm$ 6.9 & 59.03 $\pm$ 21.3 & 46.96 $\pm$ 27.8 & \textbf{85.55} $\pm$ 5.0 & 61.32 $\pm$ 23.8 \\
walker2d-medium & 77.39 $\pm$ 24.9 & \textbf{72.44} $\pm$ 18.8 & \textbf{73.19} $\pm$ 21.7 & 17.10 $\pm$ 16.9 & \textbf{72.37} $\pm$ 10.2 \\
walker2d-medium-expert & 94.81 $\pm$ 32.2 & 55.61 $\pm$ 47.8 & \textbf{73.32} $\pm$ 40.4 & 41.86 $\pm$ 34.5 & 57.86 $\pm$ 33.2 \\
\bottomrule
\textbf{MuJoCo-total} & 682.93 & \textbf{631.16} & \textbf{647.26} & 599.35 & 601.87 \\
\bottomrule
\toprule
pen-human & 1.74 $\pm$ 6.9 & -1.20 $\pm$ 2.9 & -1.46 $\pm$ 3.0 & \textbf{0.70} $\pm$ 5.2 & -0.69 $\pm$ 3.9 \\
pen-cloned & 5.10 $\pm$ 7.8 & 2.71 $\pm$ 8.2 & 3.57 $\pm$ 9.7 & 7.57 $\pm$ 8.8 & \textbf{8.13} $\pm$ 8.5 \\
door-human & -0.30 $\pm$ 0.1 & -0.34 $\pm$ 0.0 & -0.33 $\pm$ 0.0 & -0.33 $\pm$ 0.0 & -0.33 $\pm$ 0.0 \\
door-cloned & -0.34 $\pm$ 0.0 & -0.34 $\pm$ 0.0 & -0.34 $\pm$ 0.0 & -0.34 $\pm$ 0.0 & -0.34 $\pm$ 0.0 \\
hammer-human & 0.82 $\pm$ 0.4 & 0.85 $\pm$ 0.3 & 0.83 $\pm$ 0.3 & 0.91 $\pm$ 0.3 & \textbf{0.97} $\pm$ 0.4 \\
hammer-cloned & 0.26 $\pm$ 0.0 & \textbf{0.57} $\pm$ 0.6 & 0.33 $\pm$ 0.1 & 0.52 $\pm$ 0.4 & 0.46 $\pm$ 0.5 \\
relocate-human & -0.30 $\pm$ 0.0 & -0.29 $\pm$ 0.0 & -0.29 $\pm$ 0.0 & -0.30 $\pm$ 0.0 & -0.30 $\pm$ 0.0 \\
relocate-cloned & -0.30 $\pm$ 0.0 & -0.30 $\pm$ 0.0 & -0.30 $\pm$ 0.0 & -0.30 $\pm$ 0.0 & -0.30 $\pm$ 0.0 \\
\bottomrule
\textbf{Adroit-total} & 6.68 & 1.66 & 2.00 & \textbf{8.43} & 7.61 \\
\bottomrule
\toprule
antmaze-umaze-play & 73.75 $\pm$ 36.1 & 47.50 $\pm$ 37.3 & 52.75 $\pm$ 43.2 & \textbf{75.50} $\pm$ 32.0 & 51.75 $\pm$ 39.5 \\
antmaze-umaze-diverse & 12.75 $\pm$ 17.2 & 26.00 $\pm$ 26.3 & 32.25 $\pm$ 28.3 & \textbf{44.75} $\pm$ 29.5 & 25.00 $\pm$ 30.8 \\
antmaze-medium-play & 0.75 $\pm$ 2.2 & \textbf{1.00} $\pm$ 1.9 & 0.75 $\pm$ 1.8 & 0.25 $\pm$ 0.8 & \textbf{1.00} $\pm$ 2.5 \\
antmaze-medium-diverse & 0.25 $\pm$ 0.8 & 0.75 $\pm$ 2.2 & \textbf{4.00} $\pm$ 7.5 & 1.00 $\pm$ 2.4 & 2.25 $\pm$ 4.0 \\
antmaze-large-play & 0.00 $\pm$ 0.0 & \textbf{0.00} $\pm$ 0.0 & \textbf{0.00} $\pm$ 0.0 & \textbf{0.00} $\pm$ 0.0 & \textbf{0.00} $\pm$ 0.0 \\
antmaze-large-diverse & 0.00 $\pm$ 0.0 & \textbf{0.00} $\pm$ 0.0 & \textbf{0.00} $\pm$ 0.0 & \textbf{0.00} $\pm$ 0.0 & \textbf{0.00} $\pm$ 0.0 \\
\bottomrule
\textbf{Antmaze-total} & 87.50 & 75.25 & 89.75 & \textbf{121.50} & 80.00 \\
\bottomrule

    \end{tabular}
    }
    
\end{table}

\paragraph{Evaluations with TD3+BC} 
\label{para:td3+bc}
Prior to the ReBRAC, TD3+BC \citep{td3_bc} was first introduced as a minimalist solution to the core challenge of offline RL, known as \emph{extrapolation error} \citep{levine_offline_rl}. Under the general framework of behavior-regularized actor-critic approach, TD3+BC does not constrain the Q target but only regularizes the actor training objective of online TD3 by adding a behavior cloning loss. One can roughly consider TD3+BC as ReBRAC with the critic regularization coefficient being 0 and the actor regularization coefficient being 1. In addition, TD3+BC does not employ tricks involved in ReBRAC, e.g., larger networks, larger batch size, and layer normalization \citep{tarasov2023rebrac}. As a result of such a reduction to an untuned primitive ReBRAC, TD3+BC cannot effectively learn from the IRL rewards and suffers from large performance variances for \texttt{walker2d} environment and catastrophically fails on \texttt{Adroit} and \texttt{Antmaze} domains, as we show in Table \ref{tabel:td3_bc_appx}.

\begin{table}[h]
    \caption{Tuned regularization coefficients of actor and critic in ReBRAC algorithms for different reward functions. In each tuple, the first value is for the actor, and the second is for the critic.}
    \label{tabel:tuned_rebrac_params_mujoco}
    \centering
    \scalebox{0.7}
    {
    \begin{tabular}{lccccccccccccc}

\toprule
Dataset & OT & TemporalOT & Seg-match & Min-Dist \\
\hline
hopper-medium-replay & (0.01, 0.5) & (0.01, 0.5) & (0.1, 0.001) & (0.01, 0.5) \\
hopper-medium & (0.01, 0.1) & (0.01, 0) & (0.05, 0.5) & (0.01, 0.5) \\
hopper-medium-expert & (0.05, 0.1) & (0.05, 0.5) & (0.5, 0.001) & (0.1, 0) \\
halfcheetah-medium-replay & (0.001, 0.5) & (0.001, 0.5) & (0.01, 0.5) & (0.001, 0.1) \\
halfcheetah-medium & (0.001, 0.5) & (0.001, 0.5) & (0.05, 0.001) & (0.001, 0.1) \\
halfcheetah-medium-expert & (0.01, 0.001) & (0.01, 0) & (0.05, 0.5) & (0.01, 0) \\
walker2d-medium-replay & (0.1, 0.5) & (0.05, 0.5) & (0.05, 0.5) & (0.01, 0.1) \\
walker2d-medium & (0.05, 0.5) & (0.01, 0.5) & (0.05, 0.5) & (0.01, 0.1) \\
walker2d-medium-expert & (0.05, 0.1) & (0.01, 0.1) & (0.01, 0.01) & (0.01, 0.001) \\
\bottomrule

    \end{tabular}
    }
    
\end{table}
\begin{table}[h]
    \caption{Tuned regularization coefficients of actor and critic in ReBRAC algorithms for different reward functions. In each tuple, the first value is for the actor, and the second is for the critic.}
    \label{tabel:tuned_rebrac_params_antmaze}
    \centering
    \scalebox{0.7}
    {
    \begin{tabular}{lccccccccccccc}

\toprule
Dataset & OT & Seg-match\\
\hline
pen-human & (0.01, 0.001) & (0.1, 0.01) \\
pen-cloned & (0.05, 0.5) & (0.1, 0.1) \\
antmaze-umaze-play & (0.01, 0.001) & (0.01, 0.001) \\
antmaze-umaze-diverse & (0.1, 0.01) & (0.01, 0.1) \\
antmaze-medium-play & (0.001, 0.001) & (0.01, 0.001) \\
antmaze-medium-diverse & (0.002, 0.1) & (0.01, 0.1) \\
antmaze-large-play & (0.001, 0) & (0.002, 0.1) \\
antmaze-large-diverse & (0.001, 0.001) & (0.002, 0.001) \\
\bottomrule

    \end{tabular}
    }
    
\end{table}

\subsection{Multi-Expert Generalization} \label{appx: multi-demo}
\begin{figure}[ht]
    \begin{center}
       \includegraphics[width=0.8\linewidth]{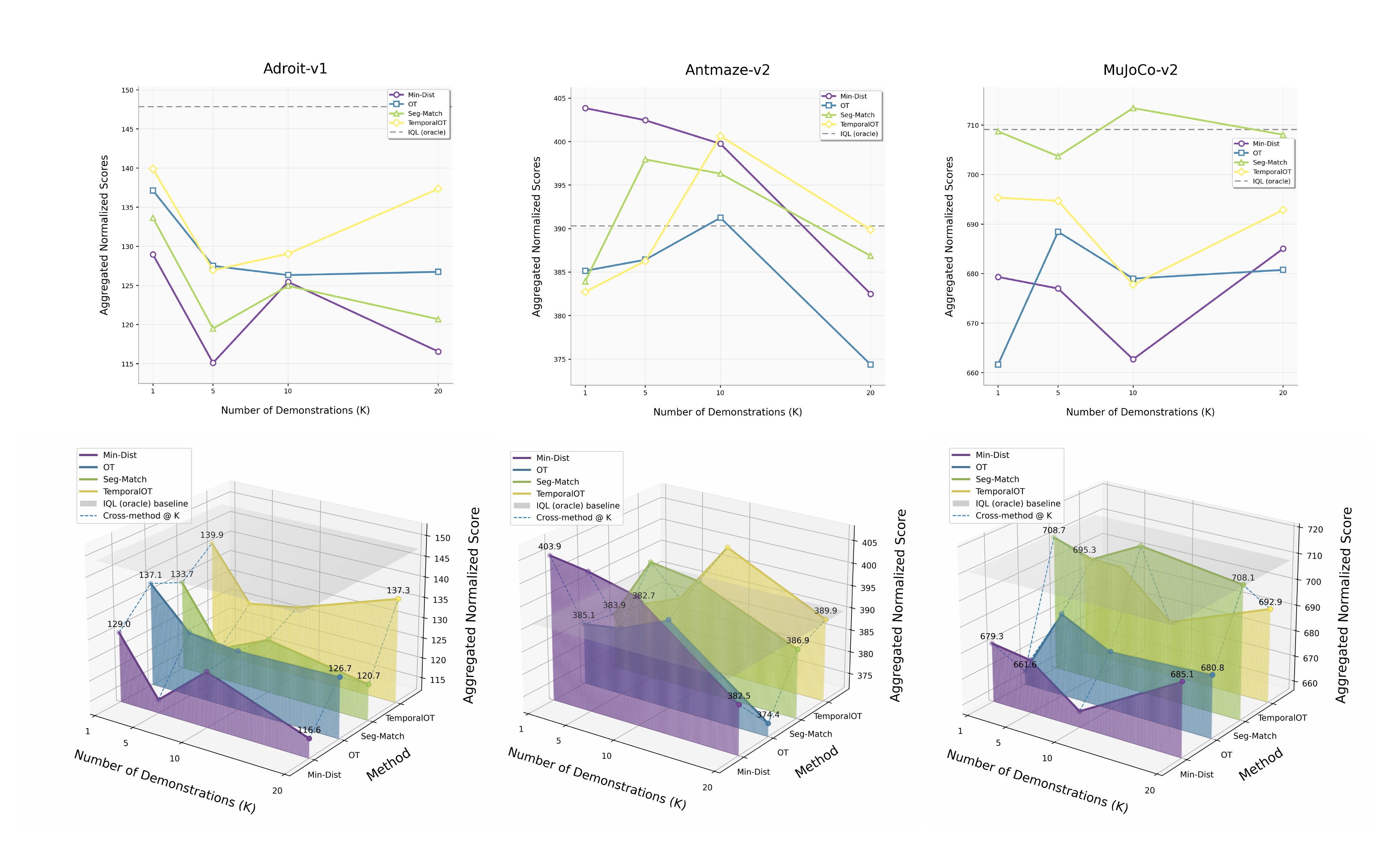}
    \end{center}
    \caption{Multi-expert ablation results for \textbf{Adroit} (left), \textbf{Antmaze} (middle) and \textbf{MuJoCo} (right).} 
    \label{fig:k-domains}
\end{figure}

In offline RL, \citet{ot_reward} selects top-K trajectories ranked by oracle return, and then fixes them across all seeds and methods. We vary $K\in\{1,5,10,20\}$ on the D4RL benchmark, using IQL as the downstream learner. Figure~\ref{fig:k-domains} reports the aggregated normalized scores for each domain, with error bars representing the accumulated standard deviations from each domain’s datasets.

\subsection{Comparison with More Related Work} \label{appx: literature}
We compare the performance of several methods discussed on MuJoCo datasets. Literature along this line of work has conducted extensive empirical analysis. Table~\ref{tabel:literature} gathers the existing results from the literature under the same problem setup, including \citet{ot_reward} and \citet{liu2023clue}.

\begin{table*}[ht]
\caption{Comparison of SQIL, UDS, IQ-Learn, ORIL, SMODICE, and CLUE on D4RL locomotion benchmarks.}
\centering
\scalebox{0.75}{
\begin{tabular}{lcccccc}
\toprule
Dataset & SQIL & UDS & IQ-Learn & ORIL & SMODICE & CLUE \\
\midrule
halfcheetah-medium         & 24.3 $\pm$ 2.7 & 42.4 $\pm$ 0.3 & 21.7 $\pm$ 1.5 & 56.8 $\pm$ 1.2 & 42.4 $\pm$ 0.6 & 45.6 $\pm$ 0.3 \\
halfcheetah-medium-replay  & 43.9 $\pm$ 1.0 & 37.9 $\pm$ 2.4& 7.7 $\pm$ 1.6  & 46.2 $\pm$ 1.1 & 38.3 $\pm$ 2.0 & 43.5 $\pm$ 0.5 \\
halfcheetah-medium-expert  & 6.7 $\pm$ 1.2  & 63.0 $\pm$ 5.7 & 2.0 $\pm$ 0.4  & 48.7 $\pm$ 2.4 & 80.9 $\pm$ 2.3 & 90.0 $\pm$ 2.4 \\
hopper-medium              & 66.9 $\pm$ 5.1 & 54.5 $\pm$ 3.0 & 29.6 $\pm$ 5.6 & 96.3 $\pm$ 0.9 & 54.8 $\pm$ 1.2 & 78.3 $\pm$ 5.4 \\
hopper-medium-replay       & 98.6 $\pm$ 0.7 & 49.3 $\pm$ 2.7 & 23.0 $\pm$ 9.4 & 56.7 $\pm$ 12.9& 30.4 $\pm$ 7.8 & 94.3 $\pm$ 6.0 \\
hopper-medium-expert       & 13.6 $\pm$ 9.6 & 53.9 $\pm$ 2.5 & 9.1 $\pm$ 2.2  & 25.1 $\pm$ 12.8& 82.4 $\pm$ 7.8 & 96.5 $\pm$ 14.7 \\
walker2d-medium            & 51.9 $\pm$ 11.7& 68.9 $\pm$ 6.2 & 5.7 $\pm$ 4.0  & 20.4 $\pm$ 13.6& 67.8 $\pm$ 6.0 & 80.7 $\pm$ 1.5 \\
walker2d-medium-replay     & 42.3 $\pm$ 5.8 & 17.7 $\pm$ 9.6 & 17.0 $\pm$ 7.6 & 71.8 $\pm$ 9.6 & 49.7 $\pm$ 4.6 & 76.3 $\pm$ 2.8 \\
walker2d-medium-expert     & 18.8 $\pm$ 13.1& 107.5 $\pm$ 1.7& 7.7 $\pm$ 2.4  & 11.6 $\pm$ 14.7& 94.8 $\pm$ 11.1& 109.3 $\pm$ 2.1 \\
\midrule
\textbf{MuJoCo-total}   & 367.0 & 495.1 & 123.5 & 433.6 & 541.5 & 714.5 \\
\bottomrule
\toprule
Dataset & BC & IQL (oracle) & OT & TemporalOT & Seg-match & Min-Dist \\
\hline
hopper-medium-replay & 29.81 $\pm$ 2.07 & 94.66 $\pm$ 8.2 & 65.59 $\pm$ 20.5 & 77.01 $\pm$ 4.9 & \textbf{87.69} $\pm$ 1.1 & \textbf{86.96} $\pm$ 2.1 \\
hopper-medium & 53.51 $\pm$ 1.76 & 63.50 $\pm$ 4.8 & \textbf{73.68} $\pm$ 4.8 & \textbf{76.30} $\pm$ 4.9 & \textbf{76.77} $\pm$ 4.5 & 72.39 $\pm$ 4.2 \\
hopper-medium-expert & 52.30 $\pm$ 4.01 & 103.67 $\pm$ 8.9 & \textbf{105.45} $\pm$ 7.6 & \textbf{104.51} $\pm$ 11.5 & \textbf{107.85} $\pm$ 4.0 & 82.67 $\pm$ 37.7 \\
halfcheetah-medium-replay & 35.66 $\pm$ 2.33 & 43.12 $\pm$ 1.6 & \textbf{41.53} $\pm$ 0.4 & \textbf{43.15} $\pm$ 0.6 & \textbf{41.03} $\pm$ 1.0 & \textbf{42.38} $\pm$ 0.9 \\
halfcheetah-medium & 42.40 $\pm$ 0.19 & 47.49 $\pm$ 0.3 & \textbf{43.46} $\pm$ 0.3 & \textbf{45.13} $\pm$ 0.2 & \textbf{43.23} $\pm$ 0.3 & \textbf{45.03} $\pm$ 0.2 \\
halfcheetah-medium-expert & 55.95 $\pm$ 7.35 & 91.24 $\pm$ 2.3 & \textbf{88.53} $\pm$ 3.4 & \textbf{89.49} $\pm$ 3.5 & \textbf{90.71} $\pm$ 2.7 & \textbf{88.72} $\pm$ 2.9 \\
walker2d-medium-replay & 21.80 $\pm$ 10.15 & 73.79 $\pm$ 8.9 & 53.87 $\pm$ 19.5 & \textbf{69.73} $\pm$ 11.4 & \textbf{70.93} $\pm$ 14.6 & \textbf{70.82} $\pm$ 8.6 \\
walker2d-medium & 63.23 $\pm$ 16.24 & 80.57 $\pm$ 4.2 & \textbf{79.58} $\pm$ 1.6 & \textbf{79.83} $\pm$ 1.8 & \textbf{80.20} $\pm$ 2.4 & \textbf{79.43} $\pm$ 1.1 \\
walker2d-medium-expert & 98.96 $\pm$ 15.98 & 111.06 $\pm$ 0.2 & \textbf{109.93} $\pm$ 0.2 & \textbf{110.19} $\pm$ 0.2 & \textbf{110.31} $\pm$ 0.2 & \textbf{110.91} $\pm$ 0.2 \\
\bottomrule
\textbf{MuJoCo-total}& 453.6 & 709.10 & 661.62 & \textbf{695.34} & \textbf{708.73} & \textbf{679.33} \\
\bottomrule

\label{tabel:literature}
\end{tabular}
}
\end{table*}

Note that simple heuristics like SQIL and UDS, which assign 0-1 rewards to the unlabeled dataset, do not work well in this challenging setup where limited expert demonstrations are available. Although CLUE seems to achieve the highest total score, we note that: First, CLUE uses an SAC expert to collect demonstrations for \emph{all} MuJoCo datasets, while other work simply uses the best episode in each dataset. Second, CLUE requires more than one expert trajectory to train representations (Table 8 \citet{liu2023clue}). Last, CLUE tunes and chooses the best temperature constant in the exponential squashing for each dataset (Table 6 \citet{liu2023clue}), while the four methods discussed in our paper stick to one value throughout.

\clearpage
\section{Online RL Evaluations}
\label{appx:metaworld_setup}

\subsection{Reward Post-processing}
Unlike the offline setting, global reward normalization based on dataset-wide return bounds is not feasible online because trajectories are collected continuously and return ranges are unknown a priori. We therefore follow the per-run reward scaling procedure used in TemporalOT \citep{temporal_ot}, applied uniformly across all reward assignment methods. 

\paragraph{Per-run scaling.}
Let $\{r_t\}$ denote the proxy rewards produced during the first training episode. We compute
\[
\text{rewards\_sum} = \sum_{t} \left| r_t \right|
\]
and define a fixed scaling coefficient
\[
\texttt{reward\_scale} = \frac{\texttt{reward\_scale\_factor}}{\text{rewards\_sum}},
\]
where we set $\texttt{reward\_scale\_factor}=10$ for all experiments. This scale is then held fixed and applied to all subsequent rewards during training. This procedure stabilizes reward magnitudes while preserving relative reward structure within each run.

\paragraph{Seg-Match smoothing.}
In MetaWorld, episodes have a fixed horizon, so Seg-Match reduces to strict timestep-to-timestep matching. To reduce high-frequency variance in the resulting reward sequence while keeping correspondence fixed, we apply a temporal Gaussian filter to the per-episode reward sequence with standard deviation $0.5$ and kernel radius $T/2$.

\paragraph{Existence of multiple experts.}
In online RL, we use the expert generation code of \citet{temporal_ot} to produce two demo trajectories, again fixed afterwards across seeds and methods. For each episode, we compute proxy rewards independently with respect to each expert demonstration and retain the reward sequence induced by the expert that yields the higher total episodic proxy return.

\subsection{Hyperparameter Settings}
\label{appx:hyperparam_online}

We adopt the same hyperparameter settings used in \citet{temporal_ot} for all methods unless otherwise stated. All policies are trained using the DrQ-v2 algorithm, with vision-based observations and a total training horizon of 1 million environment steps. Table~\ref{tab:online_hyperparams} summarizes the key parameters. Since \citet{temporal_ot} only provides expert demonstrations for \texttt{basketball-v3} and \texttt{door-open-v2} in their released code, we use their codebase and provided expert policies to collect two expert trajectories for the remaining tasks. We observed that these expert policies can produce suboptimal trajectories on some tasks, which can lead to lower absolute success rates than those reported in \citet{temporal_ot}. To keep comparisons controlled, we fix the collected expert trajectories per task and use the same two demonstrations for all reward assignment methods.

\subsection{MetaWorld Tasks}
We evaluate our methods on nine tasks from the MetaWorld benchmark \citep{metaworld2020}, following the setup in \citet{temporal_ot}. Each task involves a vision-based robotic manipulation objective with fixed episode lengths. Specifically, the \texttt{basketball-v3} and \texttt{lever-pull-v2} tasks have horizon length 175, while all other tasks are capped at 125 steps. Task descriptions and goal specifications are provided below.

\begin{itemize}
    \item \textbf{Basketball}: Grasp the ball and move it above the rim.
    \item \textbf{Button-press}: Press down a red button.
    \item \textbf{Door-lock}: Rotate the door lock knob to a target angle.
    \item \textbf{Door-open}: Open the door to a specified position.
    \item \textbf{Hand-insert}: Insert a brown block into a hole.
    \item \textbf{Lever-pull}: Pull the lever to a target height.
    \item \textbf{Push}: Push a red cylinder to a goal location on the table.
    \item \textbf{Stick-push}: Use a blue stick to push a bottle to the goal.
    \item \textbf{Window-open}: Slide the window to a designated open position.
\end{itemize}

\begin{table}[ht]
    \caption{Hyperparameters for DrQ-v2 online evaluations on MetaWorld.}
    \label{tab:online_hyperparams}
    \centering
    \scalebox{0.85}{
    \begin{tabular}{cll}
    \toprule
    & \textbf{Hyperparameter} & \textbf{Value} \\
    \midrule
    & Total environment steps & 1e6 \\
    & Evaluation frequency & Every 5e4 steps \\
    & Evaluation episodes & 10 \\
    Training & Batch size & 512 \\
    & Discount factor \( \gamma \) & 0.9 \\
    & Target network update rate \( \tau \) & 0.005 \\
    & Learning rate & 1e-4 \\
    \midrule
    & Actor hidden layers & (1024, 1024, 1024) \\
    & Critic hidden layers & (1024, 1024, 1024) \\
     & Embedding dimension & 50 \\
Network Architecture    & CNN channels & (32, 32, 32, 32) \\
    & CNN kernel sizes & (3, 3, 3, 3) \\
    & CNN strides & (2, 1, 1, 1) \\
    & CNN padding & VALID \\
    \midrule
    & Buffer size & 1.5e5 \\
DrQ-specific Settings    & Action repeat & 2 \\
     & Frame stack & 3 \\
    & Image resolution & \( 84 \times 84 \times 3 \) \\
    \midrule
    & Number of expert demonstrations & 2 \\
Reward Labeling    & Context length \( k_c \) & 3 \\
 & Temporal mask radius \( k_m \) (TemporalOT) & 10 \\

    \midrule
    & GPU model & NVIDIA A100 80GB \\
    & Parallel tasks per GPU & 5 \\
 Computation   & CPU workers per task & 2 \\
    & Memory per task & 16 GB \\
    & Approximate average execution time & 10 hours \\
    \bottomrule
    \end{tabular}
    }
\end{table}

\clearpage
\section{Further Discussion on Reward Computation}
\label{appx:generalized_reward}

\subsection{Computational Complexity} \label{appx:complexity}
We analyze the cost of computing reward labels for the various reward-labeling algorithms for a single non-expert trajectory of length \( T \). To simplify the analysis, we assume $T_e = T$. Let $d$ denote the dimension of the state. 

\paragraph{OT} The original OT method requires computing pairwise distances between all agent and expert states, forming a cost matrix of size \( T \times T \). Each entry involves a \( d \)-dimensional comparison, resulting in \( \mathcal{O}(T^2 d) \) time for cost matrix construction. Solving the entropic OT problem (e.g., via Sinkhorn iterations) typically adds no worse asymptotic cost, so the overall complexity remains \( \mathcal{O}(T^2 d) \). 

\paragraph{TemporalOT}If we ignore the cost of computing the context-aware cost matrix, the actual implementation of the temporal mask in the Temporal OT paper does not add additional computational cost on top of OT. TemporalOT sets all non-masked costs to a very large number, so that this forces the optimal coupling to assign probability mass only to expert states within the masking window. In terms of computing the context-aware cost matrix, the worst complexity is $\mathcal{O}(k_cT^2)$, where $k_c$ is the context length. Therefore, the total complexity is $\mathcal{O}((k_c+d)T^2)$.

\paragraph{Seg-Match} In contrast, the Segment-Match method assigns each agent state \( s_t \) to a single non-overlapping expert segment of fixed size, and computes its minimum distance to that segment. Since the number of comparisons per state is constant, the total cost scales as \( \mathcal{O}(T d) \). 

\paragraph{Min-Dist} The Min-Dist method requires finding, for each non-expert state, its nearest neighbor in the entire expert trajectory. A brute-force implementation incurs \( \mathcal{O}(T^2 d) \) cost. However, when \( d \) is small, we may pre-build a KD-Tree over expert states and reduce the query time to \( \mathcal{O}(\log T) \), leading to a total complexity of \( \mathcal{O}(T \log T + T d) \).

In practice, we do observe significant increases in GPU time spent on OT reward computation, as the number of expert demonstrations increases. For example, IQL with JAX implementation takes \textbf{20 minutes} for training 1 million steps on the Antmaze-large-play dataset. When the number of expert demonstrations is 20, OT with JAX implementation requires \textbf{48 minutes} for labeling this dataset. And TemporalOT with JAX needs \textbf{58 minutes}. Thus, the downstream RL algorithm is no longer the bottleneck of computation. This again emphasizes the significance of simplifying or eliminating the OT optimization.

\subsection{Generalized Simple Reward Formulation}

We describe a general framework for constructing proximity-based reward functions using temporally indexed windows over expert trajectories. The framework makes explicit the correspondence rule $\mathcal{W}(t)$ that selects which expert states are eligible for comparison at each agent timestep $t$.

Let $\tau=(s_1,\dots,s_T)$ be a non expert trajectory and $\tau^e=(s_1^e,\dots,s_{T_e}^e)$ an expert trajectory. For each timestep $t\in\{1,\dots,T\}$, define
\[
r_t \;=\; - \min_{s^e \in \mathcal{W}(t)} \mathrm{Dist}(s_t, s^e),
\]
where $\mathrm{Dist}(\cdot,\cdot)$ is a chosen distance metric (e.g., cosine or Euclidean).

\paragraph{Minimum-Distance as a window rule.}
The temporally invariant Minimum-Distance reward corresponds to the constant window
\[
\mathcal{W}(t)=\{s_1^e,\dots,s_{T_e}^e\}\quad \text{for all } t,
\]
which yields $r_t=-\min_{1\le i\le T_e}\mathrm{Dist}(s_t,s_i^e)$.

\paragraph{Segment-Matching as a window rule.}
Segment-Matching corresponds to a partition of the expert trajectory into $T$ contiguous segments and using the segment indexed by agent time. Let
\[
q=\left\lfloor \frac{T_e}{T} \right\rfloor,
\qquad
l=T_e \bmod T,
\]
and define
\[
a_t=(t-1)q + 1 + \min(t-1,l),
\qquad
b_t=tq + \min(t,l).
\]
Define the segment
\[
\Gamma_t=\left\{ s_i^e \ \middle|\ i\in[a_t,b_t]\cap\mathbb{Z} \right\}.
\]
When $T_e\ge T$, $\Gamma_t$ is nonempty for all $t$, with sizes $|\Gamma_t|\in\{q,q+1\}$. When $T>T_e$, segments become empty for $t>T_e$ under the above construction. In that case we follow the same tail rule used in the main text and set
\[
\mathcal{W}(t)=\Gamma_t \quad \text{for } t\le T_e,
\qquad
\mathcal{W}(t)=\{s_{T_e}^e\} \quad \text{for } t>T_e.
\]
This yields Segment-Matching rewards $r_t=-\min_{s^e\in\Gamma_t}\mathrm{Dist}(s_t,s^e)$ for $t\le T_e$ and $r_t=-\mathrm{Dist}(s_t,s_{T_e}^e)$ for $t>T_e$.

\paragraph{Sliding-window temporal matching.}
A simple temporal bias that interpolates between strict matching and Min-Dist is obtained by setting
\[
\mathcal{W}(t)=\{s_i^e : i\in[t-w,t+w]\cap\{1,\dots,T_e\}\}
\]
for a window radius $w\ge 0$. This provides a direct non-OT baseline with explicit temporal locality.

Overall, this formulation unifies proximity-based rewards by separating the distance metric from the correspondence rule $\mathcal{W}(t)$.

\clearpage

\section{Proof of Theorem~\ref{thm:bound}}
\label{appx:theory}

\begin{theorem}[Sufficiency Bound for Min-Dist Reward; full version] 
Let $\mathcal{M} = (\mathcal{S}, \mathcal{A}, \mathcal{P}, \mathcal{R}, \gamma)$ be a finite-horizon deterministic MDP with $\gamma \in [0,1)$, where
\begin{itemize}
    \item state-action space $\mathcal{S}\times \mathcal{A}$ is finite;
    \item transition dynamics $\mathcal{P}$ is known and deterministic;
    \item reward function $\mathcal{R}(s,a):= \underset{\tau\sim \mathcal{M}}{\mathbb{E}} \mathcal{R}_{ot}(s,a,\tau;\tau_e) \in(0,1]$ computes the stationary average OT reward over all possible trajectories $\tau\in\mathcal{M}$ containing $(s,a)\in\tau$, where $\mathcal{R}_{ot}$ is the OT reward labeling algorithm and $\tau_e$ is a fixed expert demonstration.
\end{itemize}

A reward-free offline dataset $\mathcal{D}=\{(s,a)\}$ is given, and let the corresponding behavior policy be $\pi_\beta$. Consider identical reward post-processing for different relabelings. Denote $\mathcal{D}_{ot}=\{(s,a, r_{s,a}^{ot})\}$ to be the dataset relabeled by OT rewards; and $\mathcal{D}_{min}=\{(s,a, r_{s,a}^{min})\}$ to be the dataset relabeled by Min-Dist rewards.

Assume that the optimal policy is learned through a generic constrained offline policy optimization problem: 
\[
\pi_{x}^\ast := \underset{\pi}{\argmax} \hat{J}_{\mathcal{D}_x}(\pi) - \frac{\alpha}{1-\gamma} \Div(\pi, \pi_\beta)
\] 
where $x\in\{ot, min\}$; $\hat{J}_{\mathcal{D}_x}(\pi)$ is the average return of policy $\pi$ in the empirical MDP induced by dataset $\mathcal{D}_x$; $\Div$ measures divergence between distributions; and $\alpha$ controls the constraint strength.

Following prior work \citep{kumar2020cql, yu2021conservative}, assume that (1) the labeled OT rewards $r_{s,a}^{ot}\in\mathcal{D}_{ot}$ and (2) empirical dynamics induced by the offline dataset $\mathcal{D}$ concentrate towards their mean, with high probability $\geq 1-\delta$: 
 \begin{equation*} \label{proof:assumption}
        |r_{s,a}^{ot} - \mathcal{R}(s,a)| \leq \frac{C_{R, \delta}}{\sqrt{\mathcal{D}(s,a)}}, ~~~ ||\widehat{\mathcal P}(s'|s, a) - \mathcal P(s'|s, a)||_{1} \leq \frac{C_{P, \delta}}{\sqrt{\mathcal{D}(s, a)}}.
\end{equation*}

where $C_{R, \delta}$ and $C_{P, \delta}$ are positive constants; and $\mathcal{D}(s,a)$ denotes the number of visits to $(s,a)$ in $\mathcal{D}$.

Define $\rho(s,a) = \frac{|\{(s,a,r)\in\mathcal{D}_{min}|r \neq r_{s,a}^{ot}\}|}{\mathcal{D}(s,a)}$ to be the ratio of reward mismatch between OT and Min-Dist labeling. Then, with high probability $\geq 1-\delta$, $\pi_{ot}^\ast$ is no better than $\pi_{min}^\ast$ by the following bound:
\begin{align*}
J(\pi_{ot}^\ast) - J(\pi_{min}^\ast) &\leq \frac{1}{1-\gamma} \mathbb{E}_{d_{\D}^{\pi_\beta}}\bigl[\rho(s,a) \cdot \left(\mathcal{R}(s,a)-r_{s,a}^{min}\right)\bigr] \\
& - \frac{1}{1-\gamma}\mathbb{E}_{d_{\D}^{\pi_{min}^\ast}}\bigl[\rho(s,a) \cdot \left(\mathcal{R}(s,a)-r_{s,a}^{min}\right)\bigr] \\
& - \frac{\alpha}{1-\gamma}\Div(\pi_{min}^\ast, \pi_\beta) + \text{sampling error}
\end{align*}
where $J(\pi)$ is the average return of $\pi$ in the MDP $\mathcal{M}$; and $d^{\pi}(s,a)$ denotes the marginal state-action distribution of $\pi$.
\end{theorem}

\begin{proof}
Our proof first follows the proof of Theorem 4.1 in \cite{yu2022uds}. We want to get an upper and lower bound on $J(\pi)-J(\pi^\beta)$ for any $\pi$. The major difference lies in how we define and bound the reward bias in our context.

We define the effective reward of a particular transition $(s, a, r^x_{s,a}) \in \mathcal{D}_x (x\in\{ot, min\})$ by taking into consideration contributions from both the rewards that are consistent under OT and Min-Dist relabeling and the inconsistent rewards:

\begin{align*}
    r^\mathrm{eff}_{s,a} = \rho(s,a) r^x_{s,a} + (1-\rho(s,a))r^{ot}_{s,a}
\end{align*}
where we recall that $\rho(s,a) = \frac{|\{(s,a,r)\in\mathcal{D}_{x}|r \neq r_{s,a}^{ot}\}|}{\mathcal{D}(s,a)}$ to be the ratio of reward mismatch. Then, we have the following bounds under Assumption~\ref{proof:assumption}:
\begin{align*}
    r^\mathrm{eff}_{s,a} - \mathcal{R}(s, a) 
    &= (1-\rho(s,a)) \left(r^{ot}_{s,a}- \mathcal{R}(s, a) \right) - \rho(s,a) \cdot ( \mathcal{R}(s, a)- r^x_{s,a})\\
    &\leq (1-\rho(s,a)) \frac{C_{R, \delta}}{\sqrt{\mathcal{D}(s,a)}} - \rho(s,a)(\mathcal{R}(s, a) - r^{x}_{s,a})
\end{align*}
and
\begin{align*}
   r^\mathrm{eff}_{s,a} - \mathcal{R}(s, a) 
    &= (1-\rho(s,a)) \left(r^{ot}_{s,a}- \mathcal{R}(s, a) \right) - \rho(s,a) \cdot ( \mathcal{R}(s, a)- r^x_{s,a})\\
    &\geq - (1-\rho(s,a)) \frac{C_{R, \delta}}{\sqrt{\mathcal{D}(s,a)}} - \rho(s,a)(\mathcal{R}(s, a) - r^{x}_{s,a})
\end{align*}

Then, we start to bound $J(\pi)-J(\pi^\beta)$. Note that:
\begin{align*}
    J(\pi) - J(\pi^\beta) = J(\pi) - \hat{J}(\pi)+ \hat{J}(\pi)  - \hat{J}(\pi^\beta) + \hat{J}(\pi^\beta) - {J}(\pi^\beta)
\end{align*}
and 
\begin{align}
    &\hat{J}(\pi) - J(\pi) = \frac{1}{1 - \gamma} \sum_{s, a} \left(d^\pi_{\mathcal{D}_x}(s) \pi(a|s) r^{\mathrm{eff}}_{s,a} - d^\pi(s) \pi(a|s) \mathcal{R}(s, a)\right) \nonumber \\
    &= \frac{1}{1 - \gamma} \sum_{s,a} d^\pi_{\mathcal{D}_x}(s) \pi(a|s) \left(r^{\mathrm{eff}}_{s,a} - \mathcal{R}(s, a)\right)
    + \frac{1}{1 - \gamma} \sum_{s, a}\left(d^\pi_{\mathcal{D}_x}(s) - d^\pi(s)\right) \pi(a|s) \mathcal{R}(s, a)
\end{align}

By applying the two bounds on reward bias to Equation (4), we then imitate the same derivation as Equation (8)-(16) in \cite{yu2022uds} to get the following lower bound on $J(\pi)-J(\pi^\beta)$:
\begin{align*}
J(\pi) - J(\pi_\beta) 
&= J(\pi) - \hat{J}(\pi) + \hat{J}(\pi) - \hat{J}(\pi_\beta) + \hat{J}(\pi_\beta) - J(\pi_\beta) \\
&\geq - \frac{2 \gamma C_{P,\delta}}{(1-\gamma)^2} \, \mathbb{E}_{s \sim d^\pi_{D_x}}(s) 
\left[ \frac{\sqrt{|\mathcal{A}|}}{\sqrt{|D_x(s)|}} \sqrt{\Div(\pi, \pi_\beta)(s)} + 1 \right] \\
&\quad - \frac{2 C_{R,\delta}}{1-\gamma} \, \mathbb{E}_{s,a \sim d^\pi_{D_x}} 
\left[ \frac{1-\rho(s,a)}{\sqrt{|D_x(s,a)|}} \right] \\
&\quad - \frac{1}{1-\gamma} \mathbb{E}_{s,a \sim d^{\pi_\beta}_{D_x}} 
\big[\rho(s,a) \, (\mathcal{R}(s,a)-r_x^{s,a})\big] \\
&\quad + \frac{1}{1-\gamma} \, \mathbb{E}_{s,a \sim d^\pi_{D_x}} 
\big[\rho(s,a) \cdot (\mathcal{R}(s,a)-r_x^{s,a})\big] \\
&\quad + \frac{\alpha}{1-\gamma} \Div(\pi, \pi_\beta).
\end{align*}
where $d^\pi_{D_x}$ denotes the marginal state-action distribution under an empirical MDP induced by transitions of the dataset $\mathcal{D}_x$ following policy $\pi$.

Before deriving the upper bound for $J(\pi)-J(\pi^\beta)$, note that:
\begin{align*}
\hat{J}(\pi)-\hat{J}(\pi_\beta)
&=\frac{1}{1-\gamma}\sum_{s,a}\Big(d_{\D_x}^\pi(s)\,\pi(a\mid s)-d_{\D_x}^{\pi_\beta}(s)\,\pi_\beta(a\mid s)\Big)\,r^x_{s,a} \\
&\le \frac{1}{1-\gamma}\sum_{s,a}d_{\D_x}^\pi(s)\,\pi(a\mid s)\\
&= \frac{1}{1-\gamma}
\end{align*}
where the inequality follows as $r^x_{s,a}\in(0,1]$\footnote{Note that more advanced analysis can improve this naive bound choice, but it does not eclipse the insights shed by the three terms we exhibit in the theorem statement. The upper bound on $\hat{J}(\pi^\ast_{ot})-\hat{J}(\pi_\beta)$ can be combined into the divergence term above, resulting in $(1-\alpha\text{Div}(\pi_{min}^\ast,\pi_\beta))/(1-\gamma)$, and when $\pi^\ast_{min} \neq \pi_\beta$, this term tightens the overall bound.}. Putting this into Theorem 4.1 in \cite{yu2022uds}, we get an upper bound:
\begin{align*}
J(\pi) - J(\pi_\beta) 
&= J(\pi) - \hat{J}(\pi) + \hat{J}(\pi) - \hat{J}(\pi_\beta) + \hat{J}(\pi_\beta) - J(\pi_\beta) \\
&\leq \frac{2 \gamma C_{P,\delta}}{(1-\gamma)^2} \, \mathbb{E}_{s \sim d^\pi_{D_x}}(s) 
\left[ \frac{\sqrt{|\mathcal{A}|}}{\sqrt{|D_x(s)|}} \sqrt{\Div(\pi, \pi_\beta)(s)} + 1 \right] \\
& \quad + \frac{2 C_{R,\delta}}{1-\gamma} \, \mathbb{E}_{s,a \sim d^\pi_{D_x}} 
\left[ \frac{1-\rho(s,a)}{\sqrt{|D_x(s,a)|}} \right] \\
&\quad + \frac{1}{1-\gamma} \mathbb{E}_{s,a \sim d^{\pi}_{D_x}} 
\big[\rho(s,a) \, (\mathcal{R}(s,a)-r_x^{s,a})\big] \\
&\quad - \frac{1}{1-\gamma} \, \mathbb{E}_{s,a \sim d^{\pi_\beta}_{D_x}} 
\big[\rho(s,a) \cdot (\mathcal{R}(s,a)-r_x^{s,a})\big] \\
&\quad + \frac{1}{1-\gamma}
\end{align*}

Note that:
\begin{align*}
    J(\pi_{ot}^*) - J(\pi^*_{min}) = J(\pi_{ot}^*) - J(\pi_\beta) - (J(\pi_{min}^*) - J(\pi_\beta))
\end{align*}

We want an upper bound on $J(\pi_{ot}^*) - J(\pi_\beta)$ and note that the reward mismatch $\rho(s,a)=0$ for $r_{ot}^{s,a}$:
\begin{align*}
    J(\pi_{ot}^*) - J(\pi_\beta) 
\leq \quad &\frac{2 \gamma C_{P,\delta}}{(1-\gamma)^2} \, \mathbb{E}_{s \sim d^{\pi^*_{ot}}_{D}}(s) 
\left[ \frac{\sqrt{|\mathcal{A}|}}{\sqrt{|D(s)|}} \sqrt{\Div(\pi^*_{ot}, \pi_\beta)(s)} + 1 \right] \\
& + \frac{2 C_{R,\delta}}{1-\gamma} \, \mathbb{E}_{s,a \sim d^{\pi^*_{ot}}_{D}} 
\left[ \frac{1}{\sqrt{|D(s,a)|}} \right] + \frac{1}{1-\gamma}
\end{align*}

Similarly, we want a lower bound on $J(\pi^*_{min}) - J(\pi_\beta) $:
\begin{align*}
J(\pi^*_{min}) - J(\pi_\beta) 
\geq  \quad & - \frac{2 \gamma C_{P,\delta}}{(1-\gamma)^2} \, \mathbb{E}_{s \sim d^{\pi^*_{min}}_{D}}(s) 
\left[ \frac{\sqrt{|\mathcal{A}|}}{\sqrt{|D(s)|}} \sqrt{\Div(\pi^*_{min}, \pi_\beta)(s)} + 1 \right] \\
& - \frac{2 C_{R,\delta}}{1-\gamma} \, \mathbb{E}_{s,a \sim d^{\pi^*_{min}}_{D}} 
\left[ \frac{1-\rho(s,a)}{\sqrt{|D(s,a)|}} \right] \\
&\quad - \frac{1}{1-\gamma} \mathbb{E}_{s,a \sim d^{\pi_\beta}_{D}} 
\big[\rho(s,a) \, (\mathcal{R}(s,a)-r_{min}^{s,a})\big] \\
&\quad + \frac{1}{1-\gamma} \, \mathbb{E}_{s,a \sim d^{\pi^*_{min}}_{D}} 
\big[\rho(s,a) \cdot (\mathcal{R}(s,a)-r_{min}^{s,a})\big] \\
&\quad + \frac{\alpha}{1-\gamma} \Div(\pi^*_{min}, \pi_\beta).
\end{align*}

Putting together the two inequalities, we will have the final result.

\end{proof}

\end{document}